\documentclass[review]{elsarticle}

\usepackage{amsmath}
\usepackage{amsthm}
\usepackage{amsfonts}
\usepackage{amssymb}
\usepackage{graphicx}
\usepackage[ruled,vlined]{algorithm2e}
\usepackage{enumitem}
\usepackage{bbm}
\usepackage{tabularx}
\usepackage{multirow}
\usepackage{lineno,hyperref}
\usepackage{bm}
\usepackage{xcolor}
\usepackage{threeparttable}
\usepackage[caption=false,font=footnotesize]{subfig}

\journal{Pattern Recognition}

\DeclareMathOperator*{\argmin}{argmin}
\newtheorem{theorem}{Theorem}
\newtheorem{remark}{Remark}
\newcommand{\diag}{\mathrm{diag}}
\newcommand{\R}{\mathbb{R}}
\newcommand{\Tr}{\mathrm{tr}}

\makeatletter
\newcommand{\nosemic}{\renewcommand{\@endalgocfline}{\relax}}
\newcommand{\dosemic}{\renewcommand{\@endalgocfline}{\algocf@endline}}
\let\oldnl\nl
\newcommand{\nonl}{\renewcommand{\nl}{\let\nl\oldnl}}
\makeatother

\begin{document}

\begin{frontmatter}
\title{Adaptive label thresholding methods for online multi-label classification}

\author[1]{Tingting~Zhai} \corref{cor1}   \ead{zhtt@yzu.edu.cn}
\author[1]{Hongcheng~Tang}
\author[2]{Hao~Wang} \ead{wanghao.hku@gmail.com}

\cortext[cor1]{Corresponding author}
\address[1]{College of Information Engineering, Yangzhou University, Yangzhou, China, 225127}
\address[2]{School of Computer and Software, Nanjing University of Information Science and Technology (NUIST), China, 210044}

\begin{abstract}

Existing online multi-label classification works cannot well handle the online label thresholding problem and lack the regret analysis for their online algorithms. This paper proposes a novel framework of adaptive label thresholding algorithms for online multi-label classification, with the aim to overcome the drawbacks of existing methods. The key feature of our framework is that both scoring and thresholding models are included as important components of the online multi-label classifier and are incorporated into one online optimization problem. 
Further, in order to establish the relationship between scoring and thresholding models, a novel multi-label classification loss function is derived, which measures to what an extent the multi-label classifier can distinguish between relevant labels and irrelevant ones for an incoming instance. Based on this new framework and loss function, we present a first-order linear algorithm and a second-order one, which both enjoy closed form update, but rely on different techniques for updating the multi-label classifier.
Both algorithms are proved to achieve a sub-linear regret.
Using Mercer kernels, our first-order algorithm has been extended to deal with nonlinear multi-label prediction tasks.
Experiments show the advantage of our linear and nonlinear algorithms, in terms of various multi-label performance metrics.

\end{abstract}

\end{frontmatter}

\section{Introduction}

In many real-world applications, each data instance is naturally associated with multiple semantic meanings.
For example, in an image classification system \cite{BoutellLSB04, LiSL17}, each image may contain multiple objects and thus can be described with multiple semantic classes; in the sentiment classification of microblogs \cite{LiuC15}, each microblog can be simultaneously associated with multiple emotional states; in a news classification system \cite{BurkhardtK18}, each piece of news may belong to multiple categories. Thus, owing to increasing applications, multi-label classification has attracted great attention in the past decades.

Most of research on multi-label classification, however, is focused on batch learning methods, e.g., \cite{ElisseeffW01, Lin07, HariharanZVV10, ZhangSK17, SiZKMDH17, ZhuKZ18, ChenWWG19, WangKWJ21}. These methods commonly return a real-valued predictive function as the learned model which can output a real-valued score (or confidence) on each label for an instance. Thereafter, in order to convert the scores to classification results, one \emph{label thresholding model} is additionally required.  
Various label thresholding strategies have been proposed or adopted \cite{AlotaibiF21}.
\emph{Label-wise thresholding} strategies assign a different threshold for each label and a label is predicted as relevant if its score is higher than its corresponding threshold and irrelevant otherwise \cite{Lin07, ZhangSK17}.
\emph{Instance-wise thresholding} strategies assign a different threshold for each instance and exploit the threshold to bipartition all the labels into relevant and irrelevant subsets according to their scores \cite{ElisseeffW01, FurnkranzHMB08, LargeronMG12}, or alternatively, predict directly the number of relevant labels $N_x$ for each instance and pick the top $N_x$ highest scoring labels as relevant \cite{TangRN09}.
\emph{Global thresholding} strategies adopt a fixed threshold for all labels and instances \cite{ChenWWG19, WangKWJ21}.
In the aforementioned thresholding strategies, some are tailored to specific scoring models \cite{Lin07, ZhangSK17, FurnkranzHMB08, ChenWWG19, WangKWJ21}, and some are general-purpose which can be applied to any scoring model \cite{ElisseeffW01, LargeronMG12,TangRN09}. In general-purpose strategies, the thresholding model is mostly learnt based on the output of the scoring model on all training instances and thus is closely related to the scoring model. 
Typically, once a new batch of data arrives, batch learning methods suffer from expensive re-training cost, regardless of thresholding strategies, which involves re-training of their scoring model and possibly thresholding model.
Thus, batch learning methods are not efficient for large-scale or streaming multi-labeled classification tasks.
In contrast, online learning methods can provide promising solutions for such tasks by processing data one-by-one. 
In order to learn a multi-label classifier incrementally from data in a sequential manner, online methods should no doubt be able to learn both scoring and thresholding models simultaneously in an online manner.

Although several online multi-label classification methods have been developed recently, they cannot learn the label thresholding model well. Online Sequential Multi-label Extreme Learning Machines (OSML-ELM) \cite{VenkatesanEDPW17} and ELM based Online Multi-Label Learning (ELM-OMLL) \cite{DuV20} are both ELM-based online multi-label classifiers, where ELM is a feedforward neural network with a single layer of hidden nodes. OSML-ELM uses an offline post-processing procedure to determine its global label threshold. ELM-OMLL \cite{DuV20} improves OSML-ELM by defining a new objective function based on the label ranking information and by directly fixing its global label threshold as zero in the objective function. Both ELM-based methods, however, suffer from limited performance in many cases.
Methods in \cite{ReadBHP12, NguyenNLNLS19, NguyenDLLLM19} adjust existing techniques, namely, tree-based, clustering-based and Bayesian-based methods respectively, to the online multi-label setting, and then use the label cardinality computed on a fixed or variable length window of examples as the predicted number of relevant labels.  
These thresholding strategies in \cite{ReadBHP12, NguyenNLNLS19, NguyenDLLLM19} learn the thresholding model independently of their scoring model. The potential relationship between two models is thus neglected, which loses chances to exploit the scoring model to learn an improved threshold model.
The method in \cite{OsojnikPD17} transforms the multi-label classification into a multi-target regression problem and solves the problem using a streaming multi-target regressor, and then in order to convert a multi-target prediction into a multi-label prediction, a fixed threshold is required, but how to select the threshold is left unsolved in this paper.

Some other methods are related to online multi-label learning, but aim not for online multi-label classification. 
Park and Choi applied the accelerated nonsmooth stochastic gradient descent \cite{OuyangG12} to optimize the primal form of Rank-SVM \cite{ElisseeffW01} and proposed an online label ranking method \cite{ParkC13}, which only aims to rank the relevant labels higher than irrelevant ones and thus the label thresholding for classification is disregarded.
Gong et al. \cite{GongYB20} proposed an online metric learning method based on k-Nearest Neighbor and large margin principle, but this method is not concerned about how to perform online multi-label prediction in the process of metric learning.

In this paper, we aim to address the label thresholding problem for online multi-label classification.
Our contributions are as follows:
\begin{enumerate}
\item 
A novel framework of adaptive label thresholding is proposed.
Its key feature is that both scoring and thresholding models are included as important components of our online multi-label classifier and are directly encoded into one online optimization problem so as to be learnt simultaneously in an online style.

\item 
In order to establish the relationship between scoring model and label thresholding model, a novel multi-label classification loss function is derived, which measures to what an extent the multi-label classifier can distinguish relevant labels from irrelevant ones for an incoming instance. 

\item 
Based on the framework and this loss function, we present a linear first-order algorithm and a second-order one.
In updating the multi-label classifier, the first-order algorithm uses online gradient descent, while the second-order one adopts adaptive mirror descent method. Both algorithms enjoy an efficient closed-form update and have been proved to achieve a sub-linear regret,
which reveals that, on the average, their online multi-label classifier performs as well as the best fixed multi-label classifier chosen in hindsight. 

\item
Using the kernel trick, our first-order algorithm has been extended to handle nonlinear multi-label prediction tasks. 

\item 
Experiments on nine multi-label classification datasets demonstrate the superiority of our proposed linear and nonlinear algorithms, compared with several state-of-the-art online multi-label classification algorithms, in terms of various multi-label performance metrics.
\end{enumerate}

The remaining parts of this paper are organized as follows. 
Section \ref{sec.related} reviews some related work and 
Section \ref{sec.setting} presents the notation and the problem setting.
The technical details of our proposed algorithms and their theoretical analyses are provided in Section \ref{sec.algorithm}. 
In Section \ref{sec.extension}, we extend our linear first-order algorithm so as to handle nonlinear multi-label classification tasks. Experiments are provided in Section \ref{sec.experiments}. 
Finally, Section \ref{sec.conclusion} concludes this study with some perspectives of further research.

\section{Related work}
\label{sec.related}

\textbf{Online learning}. 
Online learning algorithms represent a family of efficient and scalable machine learning techniques 
\cite{Shalev-Shwartz12, Hoi2018}.
In past decades, online algorithms for \emph{single-label} classification tasks have been well explored. 
For example, PA \cite{CrammerDKSS06}, Pegasos \cite{Shalev-ShwartzSSC11}, CW \cite{CrammerDP12}, FOGD and NOGD \cite{LuHWZL16}, WOS-ELMK \cite{DingMLCLNS18} have been developed for binary or multi-class classification tasks.
Single-label classification can be regarded as specific instances of \emph{multi-label} classification, by restricting each instance to have only one relevant label, and thus multi-label classification is generally more difficult to learn than single-label tasks due to the generalization \cite{ZhangZ07}.
Currently, several online multi-label classification algorithms have been proposed in \cite{VenkatesanEDPW17, DuV20, ReadBHP12, NguyenNLNLS19, NguyenDLLLM19, OsojnikPD17}.
These algorithms, however, cannot handle the online label thresholding problem well and lack the regret analysis.
Our algorithms are proposed to overcome the shortcomings.

\textbf{Multi-label learning}.
Existing multi-label learning methods can be categorized as \emph{problem transformation} methods and \emph{algorithm adaptation} methods \cite{ZhangZ14,GibajaV15}.
The former transform the problem of multi-label classification into the problem of binary classification, multi-class classification, or label ranking, and then solve the resulting problem using well-established methods.
The transforming techniques are often algorithm independent, typically including binary relevance \cite{BoutellLSB04, ZhangLLG18}, classifier chains \cite{ReadPHF11,TeisseyreZS19}, calibrated label ranking \cite{FurnkranzHMB08} and random $k$-labelsets \cite{TsoumakasKV11, WangKWJ21}. 
The latter modify popular techniques to handle multi-label classification problems directly.
For example, lazy learning techniques \cite{ZhangZ07, PupoMV16}, SVM \cite{ElisseeffW01}, decision trees \cite{SiZKMDH17, VensSSDB08} and neural networks \cite{ChenWWG19, BiFFK20} have been respectively adapted to the multi-label setting.
Most of the above methods are batch learning methods which cannot update their scoring models and possibly label thresholding models incrementally. In contrast, in our proposed algorithms, both scoring and thresholding models are directly encoded into one online optimization problem and thus both can be learnt in an online style.

\section{Notation and problem setting}
\label{sec.setting}

In this paper, let $[L]$ denote the set $\{1,2,\cdots,L\}$ for a positive integer $L$.
Let $\mathcal{Y} = [L]$ denote the set of all $L$ possible labels.
Let $(\bm x_1, Y_1)$, $(\bm x_2, Y_2)$, $\cdots$, $(\bm x_T, Y_T)$ be an arbitrary sequence of input examples,
where $Y_t \subseteq \mathcal{Y}$ is a label set associated with the instance $\bm x_t \in \R^d$ 
for any $t \in [T]$.
Let $\bar Y_t$ denote the complementary set of $Y_t$, that is, $\bar Y_t = \mathcal{Y} - Y_t$.
So labels in $Y_t$ are relevant to $\bm x_t$, while labels in $\bar Y_t$ are not.
Let $\mathbbm{1}[a] = 1$ if $a$ is true and $\mathbbm{1}[a] = 0$ otherwise. 
 
The online multi-label classification problem can be formulated as a repeated prediction game between the learner and its environment. 
At each online round $t$, a new data instance $\bm x_t \in \R^d$ is supplied to the learner, which is required
to predict the set of all relevant labels for $\bm x_t$ using its current model $\bm W_t$ that belongs to some known hypothesis class $\mathcal{H}$. Once the learner has submitted its prediction, say $\hat Y_t$, the true set $Y_t \subseteq \mathcal{Y}$ associated with $\bm x_t$ is revealed, and the learner suffers a loss $\ell (\bm W_t; (\bm x_t, Y_t))$ that measures its multi-label predictive performance. In light of this loss, the learner is allowed to choose a new model $\bm W_{t+1} \in \mathcal{H}$, hoping for improving its performance in the subsequent rounds.

Our goal is to devise such online multi-label classification algorithms which can achieve \emph{low regret}
relative to the best fixed multi-label classifier chosen in hindsight.
Formally, the regret that the learner suffers on a sequence of $T$ examples is defined as
\begin{equation*}
Regret_T = \sum_{t=1}^T \ell (\bm W_t; (\bm x_t, y_t)) -  \min_{\bm U \in \mathcal{H}} \sum_{t=1}^T \ell (\bm U; (\bm x_t, y_t)).
\end{equation*}
Low regret means that $Regret_T$ grows sub-linearly with the number of learning rounds, $T$, and thus implies that an online learner asymptotically matches the performance of the best fixed multi-label classifier chosen in hindsight.

\section{The proposed approach}
\label{sec.algorithm}

\subsection{The framework of adaptive label thresholding}

We first use linear approaches to solve the online multi-label classification problem.
At each online round $t$, our learner maintains a multi-label classifier $\bm W_t = [\bm w_t^{(1)}, \cdots, \bm w_t^{(L)}, \bm w_t^{(L+1)}] \in \R^{d \times (L+1)}$, which consists of $L$ label predictors for computing scores for each label and one additional threshold predictor $\bm w_t^{(L+1)}$ for determining the label threshold.
When a new instance $\bm x_t \in \R^d$ arrives, our learner uses $\bm W_t$ to predict the relevant labels for $\bm x_t$. Specifically, a real-valued score $\bm x_t ^\top \bm w_t^{(i)}$ is assigned to each label $i \in \mathcal{Y}$
and a threshold score $\bm x_t ^\top \bm w_t^{(L+1)}$ is also computed.
Then the predicted set of relevant labels at round $t$ is obtained as 
$\hat Y_t = \{i \in \mathcal{Y}:  \bm x_t ^\top \bm w_t^{(i)} > \bm x_t ^\top \bm w_t^{(L+1)}\} $.
Namely, labels with higher scores than the threshold score are regarded as relevant to $\bm x_t$, 
while labels with lower scores than the threshold score are predicted as irrelevant.
When the true set $Y_t$ of $\bm x_t$ is revealed, our learner improves $\bm W_t$ to a new classifier $\bm W_{t+1}$.
According to the above descriptions, we summarize the framework of Adaptive Label Thresholding (ALT) in Algorithm \ref{alg1}.

Different from existing approaches, our framework explicitly learns both scoring model and thresholding model simultaneously in the online learning process. Based on this new framework, we construct two algorithms, for which, the main difference lies in
how to update $\bm W_t$ to $\bm W_{t+1}$. 

\begin{algorithm} 
	\LinesNumbered
	\caption{The Framework of Adaptive Label Thresholding (ALT)}
	\label{alg1}
	\nosemic
	\KwIn{Hyperparameters}
	\KwOut{$\bm W_{T+1}$}
	$\bm{W}_1 = [\bm w_1^{(1)}, \cdots, \bm w_1^{(L+1)}] = [\mathbf{0}, \cdots, \mathbf{0}] $\;
	\For {$t=1, 2, \cdots, T$}
	{
	Observe $\bm{x}_t \in \R^d$\; 
	Predict the set of relevant labels 
	$\hat{Y}_t = \{ i \in \mathcal{Y}: \bm x_t^\top \bm w_t^{(i)} > \bm x_t^\top \bm w_t^{(L+1)} \}$\;
	Receive the true set of relevant labels $Y_t \subseteq \mathcal{Y}$\;
	Update $\bm W_t$ to $\bm W_{t+1}$ \; \label{update} 
	}
\end{algorithm}

\subsection{First-order ALT algorithm (FALT)}

At the end of round $t$, we expect the online learner to find a new classifier that solves the following problem:
\begin{align} 
&\bm W_{t+1} = \argmin_{\bm W \in \R^{d \times (L+1)}} 
\frac{1}{2}  || \bm W - \bm W_t ||_{F}^2     \nonumber \\
&\mbox{ s.t. \qquad}
\bm x_t ^\top \bm w^{(i)} - \bm x_t ^\top \bm w^{(L+1)} \geq 1 ,  \ \forall i  \in Y_t  \nonumber \\
&\mbox{\qquad\qquad}
\bm x_t ^\top \bm w^{(L+1)} - \bm x_t ^\top \bm w^{(j)} \geq 1 , \ \forall j \in \bar Y_t  
\label{hard}
\end{align}
where $\bm W = [\bm w^{(1)}, \cdots, \bm w^{(L+1)}]$ and $||\cdot||_{F}$ denotes the Frobenius norm of a matrix.
The consideration for this update is two-fold. 
On the one hand, $\bm W_{t+1}$ is kept close to $\bm W_{t}$ in order to retain the information learnt from previous rounds.
On the other hand, the update requires $\bm W_{t+1}$ to not only correctly separate relevant labels from irrelevant ones for the instance $\bm x_t$, but also separate with high confidence.
By imposing such constraints, relations between the predictors of relevant labels and the label threshold predictor and relations between the label threshold predictor and the predictors of irrelevant labels can both be explicitly considered by our multi-label classifier.

In presence of label noise and outliers, forcing aggressively the new classifier $\bm W_{t+1}$ to satisfy all the $L$ hard-margin constraints in problem (\ref{hard}) may cause $\bm W_{t+1}$ to update dramatically towards a wrong direction and thus bring undesirable consequences.
To cope with such problem, the common technique for deriving soft-margin classifiers \cite{Vapnik1998, CrammerDKSS06} is adopted and nonnegative slack-variables are introduced into problem (\ref{hard}) to relax the $L$ constraints,
which leads to the following optimization problem:
\begin{align} 
 (\bm W_{t+1}, \bm {\xi}_{t+1}) 
 &=  \argmin_{\bm W, \bm \xi} \left\{ \frac{1}{2} ||\bm W  - \bm W_t||_F^2 
 + \eta \bigl( \frac{1}{|Y_t|} \sum_{i \in Y_t} \xi_i + \frac{1}{|\bar Y_t|} \sum_{i \in \bar Y_t} \xi_i  \bigr)  \right\} \nonumber \\
&\mbox{ s.t. \qquad}
\bm x_t ^\top \bm w^{(i)} - \bm x_t ^\top \bm w^{(L+1)} \geq 1 - \xi_i ,  \ \forall i  \in Y_t  \nonumber \\
&\mbox{\qquad\qquad}
\bm x_t ^\top \bm w^{(L+1)} - \bm x_t ^\top \bm w^{(j)} \geq 1 - \xi_j , \ \forall j \in \bar Y_t \nonumber \\
&\mbox{\qquad\qquad}    
\xi_i \geq 0, \ \forall i \in \{1, 2, \cdots, L\}
\label{soft}
\end{align}
where $\eta > 0$ is used to control the tradeoff between the first regularization term and the second slack variable term.
It is worth noting that the weighted sum of slack variables, instead of their unweighted sum, is used in the optimization objective, 
in order to avoid the situation that one sum dominates over the other one in 
$\sum_{i \in Y_t} \xi_i$ and $\sum_{i \in \bar Y_t} \xi_i$.
  
Further, let $\xi_i \!=\! \max \left\{0,  1 \!-\! (\bm x_t ^\top \bm w^{(i)} \!-\! \bm x_t ^\top \bm w^{(L+1)})\right\}$ 
for any $i \in Y_t$, 
and let $\xi_j \!=\! \max \left\{0,  1 \!-\! (\bm x_t ^\top \bm w^{(L+1)} \!-\! \bm x_t ^\top \bm w^{(i)})\right\}$
for any $j \in \bar Y_t$. 
Then, all constraints in (\ref{soft}) can be eliminated, so that 
the constrained problem (\ref{soft}) is transformed into the following unconstrained one:
\begin{equation} 
\label{opt}
\bm W_{t+1} = \argmin_{\bm W} 
\frac{1}{2 \eta}  || \bm W - \bm W_t ||_F^2  +  f_t (\bm W)
\end{equation}
where $f_t (\bm W)$ is defined as
\begin{align}
f_t (\bm W) &\!=\!  
\frac{1}{|Y_t|} \sum_{i \in Y_t} \max \left\{0,  1 \!-\! (\bm x_t ^\top \bm w^{(i)} \!-\! \bm x_t ^\top \bm w^{(L+1)})\right\} \nonumber\\
&\!+\! \frac{1}{|\bar Y_t|} \sum_{i \in \bar Y_t} \max \left\{0,  1 \!-\! (\bm x_t ^\top \bm w^{(L+1)} \!-\! \bm x_t ^\top \bm w^{(i)})\right\}.
\label{loss}
\end{align}

It is difficult to solve problem (\ref{opt}) directly due to the piecewise linear property of $f_t(\bm W)$. 
Instead of directly optimizing $f_t(\bm W)$, we resort to its first-order approximation, that is,
$f_t(\bm W) \approx  f_t(\bm W_t) + \sum_{i=1}^{L+1}  (\nabla_t^{(i)})^\top (\bm w^{(i)} -\bm w_t^{(i)})$,
where $\nabla_t^{(i)} = \nabla_{\bm w^{(i)}} f_t(\bm W_t)$.
By replacing $f_t(\bm W)$ in (\ref{opt}) with its first-order approximation and eliminating the terms 
irrelevant to $\bm W$, problem (\ref{opt}) is simplified as follows: 
\begin{equation} 
\label{final}
\bm W_{t+1} \!=\! \argmin_{\bm W} 
\frac{1}{2 \eta}  || \bm W - \bm W_t ||_F^2  + \sum_{i=1}^{L+1}  (\nabla_t^{(i)})^\top \bm w^{(i)}
\end{equation}

The benefit of this simplification is that an efficient closed-form solution can be derived, 
meanwhile, such update can achieve a sub-linear regret.
Since the objective function in (\ref{final}) is differentiable and convex with respect to each $\bm w^{(i)}$,
the necessary and sufficient condition for $\bm w_{t+1}^{(i)}$ to be optimal is
the gradient of the objective function with respect to $\bm w^{(i)}$ at $\bm w_{t+1}^{(i)}$ is zero vector, that is,
$\frac{1}{\eta} (\bm w_{t+1}^{(i)} - \bm w_{t}^{(i)}) + \nabla_t^{(i)} = \bm 0 $.
Solving this equation, we get the following update:
\begin{equation} 
\label{rule}
\bm w_{t+1}^{(i)} = \bm w_{t}^{(i)} - \eta \nabla_t^{(i)},  \quad \forall i \in [L+1].
\end{equation}
where
\begin{equation*}
\nabla_t^{(i)} \!=\!
\begin{cases}
-\frac{a_t^{(i)}}{|Y_t|} \bm x_t,  		&\mbox { if } i \in Y_t \\
 \frac{b_t^{(i)}}{|\bar Y_t|} \bm x_t,  &\mbox { if } i \in \bar Y_t \\ 
 (\frac{a_t}{|Y_t|}  -\frac{b_t}{|\bar Y_t|} ) \bm x_t,  & \mbox{ if } i = L+1
\end{cases}
\end{equation*}
with $a_t^{(i)} = \mathbbm{1} [\bm x_t^\top \bm w_t^{(i)} \!-\! \bm x_t^\top \bm w_t^{(L+1)} < 1]$,
$b_t^{(i)} = \mathbbm{1} [\bm x_t^\top \bm w_t^{(L+1)} \!-\! \bm x_t^\top \bm w_t^{(i)} < 1]$,
$a_t = \sum_{j \in Y_t} a_t^{(j)}$ and $ b_t = \sum_{j \in \bar Y_t} b_t^{(j)}$. 

Algorithm \ref{alg1} using (\ref{rule}) for computing the new multi-label classifier $\bm W_{t+1}$ is termed 
``First-order ALT (FALT)", since only the gradient of the loss function $f_t(\bm W)$ is required.

We now analyze the regret of FALT relative to the best fixed multi-label classifier $\bm U_* = [\bm u_*^{(1)}, \cdots, \bm u_*^{(L+1)}] \in \R^{d \times (L+1)}$ on an arbitrary sequence of examples,
where $\bm U_* $ is chosen in hindsight and has the same predictive form $\hat Y_t = \{i \in \mathcal{Y}:  \bm x_t ^\top \bm u_*^{(i)} > \bm x_t ^\top \bm u_*^{(L+1)}\}$ as our classifier.

\begin{theorem}
\label{theorem1}
Let $(\bm x_1, Y_1), \cdots, (\bm x_T, Y_T) $ be an arbitrary sequence of input examples, where $\bm x_t \in \R^d$, $Y_t \subseteq \mathcal{Y}$, and $||\bm x_t|| \leq R$ for all $t$. If FALT is run on this sequence of examples, then, for any $\bm U_* \in \argmin_{\bm U \in \R^{d \times (L+1)}} \sum_{t=1}^T f_t (\bm U)$, it holds that
\begin{align}
\label{regret}
\sum_{t=1}^T \bigl( f_t (\bm W_t) - f_t (\bm U_*) \bigr) 
\leq \frac{1}{2\eta} ||\bm U_*||_{F}^2 + \eta R^2 T
\end{align}
Setting $\eta = {||\bm U_*||_{F}}/{(R \sqrt{2 T})}$, then the regret becomes
\begin{align}
\label{bound}
\sum_{t=1}^T \bigl( f_t (\bm W_t) - f_t (\bm U_*) \bigr) \leq \sqrt{2} R ||\bm U_*||_{F} \sqrt{T} .
\end{align}
\end{theorem}

\begin{remark}
The regret bound (\ref{bound}) is sub-linear, which implies that on the average, FALT performs as well as $\bm U_*$.
Moreover, one advantage of this bound is that it is independent of $L$, the number of labels.
\end{remark}

\subsection{Second-order ALT algorithm (SALT)}

Essentially, FALT relies on $f_t(\bm W)$ defined in (\ref{loss}) to evaluate its multi-label predictive performance 
at online round $t$ and exploits online gradient descent to minimize this loss.
Thus, each element in $\bm W_t$ is updated in the same learning rate $\eta$.
In many classification tasks where data instances are very sparse and only have a few non-zero features, methods that can update each dimension of the classifier with an adaptive learning rate are deemed better than online gradient descent \cite{DuchiHS11}.
Therefore, in this section, we present a novel second-order ALT algorithm where $f_t(\bm W)$ is minimized using Adaptive Mirror Descent method (AMD) \cite{DuchiHS11} which can endow each element in $\bm W_{t}$ with an adaptive learning rate by constructing an approximation to the Hessian of $f_t(\bm W)$. 
  
Let $\ell_t(\bm w_t)$ be an instantaneous convex loss for measuring the predictive inaccuracy of an online binary classifier $\bm w_t \in \R^d$. The sub-differential set of $\ell_t$ at the point $\bm w$ is denoted by $\partial \ell_t(\bm w)$.
We use $\bm I$ as an identity matrix.
For a vector $\bm s_t$, its $i$th element is denoted by $s_{t,i}$ and we use $\diag(\bm s_t)$ to denote a diagonal matrix with the elements in $\bm s_t$ on the diagonal line. 
For a $d\times t$ matrix $\bm G_{1:t}$, its $i$th row vector is denoted by $\bm G_{1:t,i}$.
For any $p \in [1, \infty]$, the $\ell_p$ norm of a vector $\bm w$ is denoted by $\|\bm w\|_p$, and 
the Mahalanobis norm of $\bm w$ for a positive definite matrix $\bm H$ is denoted by $\|\bm w\|_{\bm H}$
which is given by $\sqrt{\bm w ^\top \bm H \bm w}$. 
Based on the notation, the pseudocode of AMD for binary classification is summarized in Algorithm \ref{alg2}.

\begin{algorithm} 
	\LinesNumbered
	\caption{AMD \cite{DuchiHS11}}
	\label{alg2}
	\nosemic
	\KwIn{$\delta > 0$, $\eta > 0$}
	\KwOut{$\bm w_{T+1} \in \R^d$}
	 $\bm{w}_1 = \mathbf{0}$\;
	\For {$t=1, 2, \cdots, T$}
	{
	Observe $\bm{x}_t \in \R^d$ \; 
	Predict the label $\hat y_t = \textrm{sign} (\bm w_t^\top \bm x_t)$\;
	Receive the true label $y_t \in \{-1, +1\}$\;
	Let $\bm g_t \!\in\! \partial \ell_t(\bm w_t)$ and $\bm{G}_{1:t} \!=\! [\bm{g}_{1},\cdots,\bm{g}_t] \in \R^{d \times t}$\;
	Let $\bm H_t = \delta \bm{I}\!+\! \diag(\bm{s}_t)$ where $s_{t,i} = ||\bm{G}_{1:t,i} ||_2$\;
	$\bm{w}_{t+1} = \argmin_{\bm{w} \in \R^d}
	\left\{\bm{g}_t^\top \bm{w} + \frac{1}{2\eta} \| \bm w - \bm w_t \|_{\bm H_t}^2 \right\}$\;
	}
\end{algorithm}

In order to apply AMD to the multi-label classification tasks, we first replace the binary classifier $\bm w_t$ with our multi-label classifier $\bm W_t$ and change the predictive rule to the one used in Algorithm \ref{alg1}, and then specify the loss function as $f_t(\bm W)$ in (\ref{loss}). We next need to calculate the subgradient of $f_t(\bm W)$ with respect to $\bm W$ at $\bm W_t$, which can be transformed into calculating the subgradient of $f_t(\bm W)$ with respect to each $\bm w^{(i)}$ at $\bm W_t$, that is, calculating $\nabla_t^{(i)}$ for any $i \in [L+1]$.
To understand the transformation, one can image the matrix $\bm W_t$ as a long column vector whose elements are taken columnwise from the matrix.
After that, following the same calculation process as Algorithm \ref{alg2}, we obtain the proposed Second-order ALT (SALT) in Algorithm \ref{alg3}, where the problem at Step \ref{secondorder} has a closed-form solution:
$$\bm w_{t+1}^{(i)} = \bm w_{t}^{(i)} - \eta (\bm H_t^{(i)})^{-1} \nabla_t^{(i)}, \quad \forall i \in [L+1].$$
In light of $\bm H_t^{(i)}$, all elements in $\bm W_{t+1}$ are updated in different learning rates.

\begin{algorithm} 
	\LinesNumbered
	\caption{Second-order ALT (SALT)}
	\label{alg3}
	\nosemic
	\KwIn{$\delta>0$, $\eta > 0$}
	\KwOut{$\bm W_{T+1}$}
	$\bm{W}_1 = [\bm w_1^{(1)}, \cdots, \bm w_1^{(L+1)}] = [\mathbf{0}, \cdots, \mathbf{0}] $\;
	\For {$t=1, 2, \cdots, T$}
	{
	Observe $\bm{x}_t \in \R^d$\; 
	Predict the set of relevant labels 
	$\hat{Y}_t = \{ i \in \mathcal{Y}: \bm x_t^\top \bm w_t^{(i)} > \bm x_t^\top \bm w_t^{(L+1)} \}$\;
	Receive the true set of relevant labels $Y_t \subseteq \mathcal{Y}$\;
	$\forall i \in [L+1]$, get $\nabla_t^{(i)}$ and then set
	$\bm{G}_{1:t}^{(i)} \!=\! [\nabla_1^{(i)},\cdots,\nabla_t^{(i)}] \in \R^{d \times t}$\; 
	$\forall i \in [L+1]$, $\bm H_t^{(i)} = \delta \bm{I}\!+\! \diag(\bm{s}_t^{(i)})$ 
	where $\forall j \in [d]$, $s_{t,j}^{(i)} = ||\bm{G}_{1:t,j}^{(i)} ||_2$\;
	$\bm W_{t+1} = \argmin_{\bm W}\!\sum_{i=1}^{L+1} \!\left(\! \frac{1}{2\eta} ||\bm w^{(i)} \!-\! \bm w_t^{(i)}||^2_{\bm H_t^{(i)}} \!+\! (\nabla_t^{(i)})^\top\! \bm w^{(i)} \!\right)\!$ \label{secondorder}
	}
\end{algorithm}

Now we start to analyze the regret of SALT relative to the best fixed multi-label classifier
$\bm U_* \in \R^{d \times (L+1)}$ chosen in hindsight.

\begin{theorem}
\label{theorem2}
Let $(\bm x_1, Y_1), \cdots, (\bm x_T, Y_T) $ be an arbitrary sequence of input examples, 
where $\bm x_t \in \R^d$ and $Y_t \subseteq \mathcal{Y}$ for all $t$.
If SALT is run on this sequence of examples, then, for any 
$\bm U_* \in \argmin_{\bm U \in \R^{d \times (L+1)}} \sum_{t=1}^T f_t (\bm U)$, it holds that
\begin{align}
\label{regret1}
\sum_{t=1}^T \bigl( f_t (\bm W_t) \!-\! f_t (\bm U_*) \bigr) 
\leq (\frac{Q}{2 \eta} + \eta)  \sum_{i=1}^{L+1} \sum_{j=1}^d \| \bm G_{1: T, j}^{(i)}\|_2
+ \frac{\delta}{2 \eta} \|\bm U_*\|_F^2
\end{align}
where $Q = {\max_{i \in [L+1], t \in [T]} \|\bm w_t^{(i)} - \bm u_*^{(i)}\|_{\infty}^2}$.
\end{theorem}

\begin{remark}
According to \cite{DuchiHS11}, the bound (\ref{regret1}) is sublinear and
when the gradient vectors are sparse, SALT is expected to perform better. 
\end{remark}

\section{Extension of FALT to nonlinear multi-label classification tasks}
\label{sec.extension}

So far, we focus on the linear multi-label classifier 
$\bm W = [\bm w^{(1)},\cdots, \bm w^{(L+1)}] \in \R^{d \times (L+1)}$ with the predictive form of $\hat Y_t = \{i \in \mathcal{Y}:  \bm x_t ^\top \bm w^{(i)} > \bm x_t ^\top \bm w^{(L+1)}\}$. 
It is noteworthy that FALT can be easily generalized using the kernel trick. 
Indeed, each component classifier of $\bm W_t$ can be expressed by
$\bm w_{t}^{(i)}  = \sum_{k=1}^{t-1} \tau_k^{(i)} \bm x_k$ for any $i \in [L+1]$
where 
\begin{equation*}
 \tau_k^{(i)} =
\begin{cases}
\eta ( \frac{\mathbbm{1}[i \in Y_k] a_k^{(i)}}{|Y_k|} - \frac{\mathbbm{1}[i \in \bar Y_k] b_k^{(i)}}{|\bar Y_k|} ), 
 &\mbox{ if } i \in [L] \\
\eta ( \frac{b_k}{|\bar Y_k|} - \frac{a_k}{|Y_k|} ),
&\mbox{ if } i = L+1.
\end{cases}
\end{equation*}

By introducing a Mercer kernel $\mathcal{K}(\cdot)$ that induces a nonlinear feature mapping $\phi(\cdot)$, 
we can get a nonlinear component classifier $\bm w_{t}^{(i)} = \sum_{k=1}^{t-1} \alpha_k^{(i)} \phi(\bm x_k)$, 
where $\alpha_k^{(i)}$ is obtained by replacing $\bm x_k$ occurring in $\tau_k^{(i)} $ with $\phi (\bm x_k)$.
Exploiting the kernel trick, scores for all labels and the threshold can be efficiently computed:
\begin{equation*}
\phi(\bm x_t)^\top \bm w_{t}^{(i)} 
=  \sum_{k=1}^{t-1} \alpha_k^{(i)} \phi(\bm x_k)^\top \phi(\bm x_t)
=  \sum_{k=1}^{t-1} \alpha_k^{(i)} \mathcal{K}(\bm x_k, \bm x_t) .
\end{equation*}
Therefore, it is not necessary to explicitly compute $\phi(\cdot)$, but instead nonlinear multi-label prediction is achieved by replacing the inner product in the original feature space with a simple kernel function operation. 
According to \cite{Vapnik1998}, Theorem \ref{theorem1} provided for linear FALT also holds for kernelized FALT.

By contrast, it is not easy to extend SALT using the kernel trick
since computing the matrix $\bm H_t^{(i)}$ requires us to explicitly compute $\phi(\cdot)$.

\section{Experiments}
\label{sec.experiments}

\subsection{Datasets}
The datasets used are presented in Table \ref{data}, which are selected to cover various domains and present diverse characteristics. The first three datasets can be downloaded from LIBSVM website\footnote{\url{https://www.csie.ntu.edu.tw/~cjlin/libsvmtools/datasets/}} and the remaining ones are available at Mulan website\footnote{\url{http://mulan.sourceforge.net/datasets-mlc.html}}.

\begin{table*}[!htp]
	\renewcommand{\arraystretch}{1.2}
	\caption{A summary of datasets in the experiments} 
	\label{data}
	\footnotesize
	\centering \tabcolsep =1pt
	\begin{threeparttable}
	\begin{tabular}{lccccccc}
	\hline
	 Dataset        & domain     &\#training   &\#testing  &\#features  &\#labels  & card  &fea.density  \\
	\hline
	Rcv1v2(industries) &text      & 23,149         & 781,265        & 47,236      &313     &0.696   &0.0016        \\
	Rcv1v2(regions)    & text      & 23,149         & 781,265        & 47,236      &228     &1.312     &0.0016      \\
	Rcv1v2(topics)      & text      & 23,149         & 781,265        & 47,236      &101     &3.241    &0.0016       \\
       Bibtex                 & text       &4,880           & 2,515          & 1,836         & 159     &2.402    &0.0374     \\
       Birds                   & audio    & 322             & 323             & 260           & 19      &1.014    &0.5858     \\
       Scene                 & images   &1,211            & 1,196           & 294          & 6        &1.074    &0.9885     \\
       Emotions             & music      &391            & 202             &72             & 6      &1.868    &0.9952    \\
       Yeast                  & biology     & 1,500           & 917              &103            & 14     &4.237    &1.0000    \\
       Mediamill 		    & video      & 30,993         & 12,914          & 120           &101      &4.376    &1.0000      \\
	\hline  
	\end{tabular}
	\begin{tablenotes}
	\footnotesize
	\item [1]  ``card" represents the label cardinality, which equals to $\frac{1}{M} \sum_{i=1}^M |Y_i|$ 
	where $M$ =  \# training + \# testing. 
	\item [2] fea.density = $\frac{1}{M \cdot d} \sum_{i=1}^M \mathsf{nnz}(\bm x_i)$ where  $d$ = \# features and 
	 $\mathsf{nnz}(\cdot)$ counts the number of non-zero elements in one vector.
	\end{tablenotes}
	\end{threeparttable}
\end{table*}

\subsection{Performance metrics}
Owing to the complexity of the multi-label classification, it is suggested to
compare algorithms from different perspectives \cite{ZhangZ14}. 
Thus, the following metrics are used.
\begin{itemize}
\item Precision (Psn), Recall (Rcal) and F1-measure (F1): 
$\textrm{Psn} \!=\! \frac{1}{N} \sum_{i=1}^N \frac{|Y_i \cap \hat Y_i |}{|\hat Y_i|}$,
$\textrm{Rcal} \!=\! \frac{1}{N} \sum_{i=1}^N \frac{|Y_i \cap \hat Y_i |}{|Y_i|}$, and
$\textrm{F1} = \frac{2*\textrm{Psn}*\textrm{Rcal}}{\textrm{Psn} + \textrm{Rcal}}$
where $N$ is the number of examples for evaluation.

\item MacroF1 and MicroF1:
Let $tp_j, fp_j$ and $fn_j$ denote the number of \emph{true positive, false positive}
and \emph{false negative} testing examples with respect to the label $j$. Then,
$\textrm{MacroF1} \!=\! \frac{1}{L} \sum_{j=1}^L \frac{2*tp_j}{2*tp_j + fp_j + fn_j}$ and
$\textrm{MicroF1} \!=\! \frac{2* \sum_{j=1}^L tp_j}{\sum_{j=1}^L (2*tp_j + fp_j + fn_j)}$.
Note that both are label-based metrics, which evaluate the performance 
on each label respectively and return the result across all labels.

\item Hamming loss (Hl): 
Let $\Delta$ stands for symmetric difference between two sets, then
$\textrm{Hl} \!=\! \frac{1}{N * L} \sum_{i=1}^N  | \hat Y_i  \Delta Y_i |$.

\item Ranking loss (Rl): 
Let $h(\bm x_i, j)$ denote the real-valued score assigned to the label $j$ for the instance $\bm x_i$, 
then
$\textrm{Rl} \!=\! \frac{1}{N} \! \sum_{i=1}^N \!
\frac{\sum_{j \in Y_i, k \in \bar Y_i} \!
\mathbbm{1} [h(\bm x_i, j)  \leq h(\bm x_i, k)]}{|Y_i| * |\bar Y_i|}$.
\end{itemize}

For Hamming loss and Ranking loss, the smaller the metric values, the better the method's performance.
For the other metrics, the larger the metric values, the better the method's performance.

\subsection{Baselines}
To demonstrate the advantage of our proposed algorithms, 
we have compared them with the following online multi-label classification algorithms:
\begin{itemize}
\item OSML-ELM \cite{VenkatesanEDPW17}: Online Sequential Multi-Label Extreme Learning Machine.
\item ELM-OMLL \cite{DuV20}: Extreme Learning Machine based Online Multi-Label Learning.
\item PA-I-BR(l) and PA-II-BR(l):    
Two Passive-Aggressive (PA) algorithms for linear binary classification, namely, PA-I and PA-II \cite{CrammerDKSS06}, were implemented and adapted to the multi-label classification setting using Binary Relevance (BR) \cite{ZhangZ14}. 
\item PA-I-BR(k) and PA-II-BR(k): the kernelized extension of PA-I-BR(l) and PA-II-BR(l), respectively.
\item FALT(l): Our proposed linear first-order algorithm.
\item FALT(k): The kernelized extension of FALT(l).
\item SALT: Our proposed linear second-order algorithm.
\end{itemize}

All kernelized algorithms use the RBF kernel with hyperparameter $\delta^2$, that is, 
$\mathcal{K} (\bm x_i ,\bm x_j) = \exp{(- \frac{||\bm x_i -\bm x_j ||^2}{2 \delta^2})}$,
and the kernel matrix is precalculated for accelerating the computation.
It is reasonable to compare linear algorithms on linear classification tasks and compare nonlinear algorithms on nonlinear tasks.
So we evaluate the linear PA-I-BR(l), PA-II-BR(l), FALT(l) and SALT, only on the first four text multi-label classification datasets in Table \ref{data}, and evaluate PA-I-BR(k), PA-II-BR(k) and FALT(k) on the other datasets.

\subsection{Sensitivity analysis}
\label{sec.paraexperiments}

We first analyze the hyperparameter sensitivity of our FALT(l) and SALT.
Except the intrinsic hyperparameters of each algorithm, we also examine whether multiple learning on each example can significantly improve various performance metrics. Thus, an additional hyperparameter $M$ that defines the maximum number of learning times on each example at each online round is considered.

\begin{figure*}[htp]
	\centering
	\subfloat[]{\includegraphics[width=0.33 \textwidth]{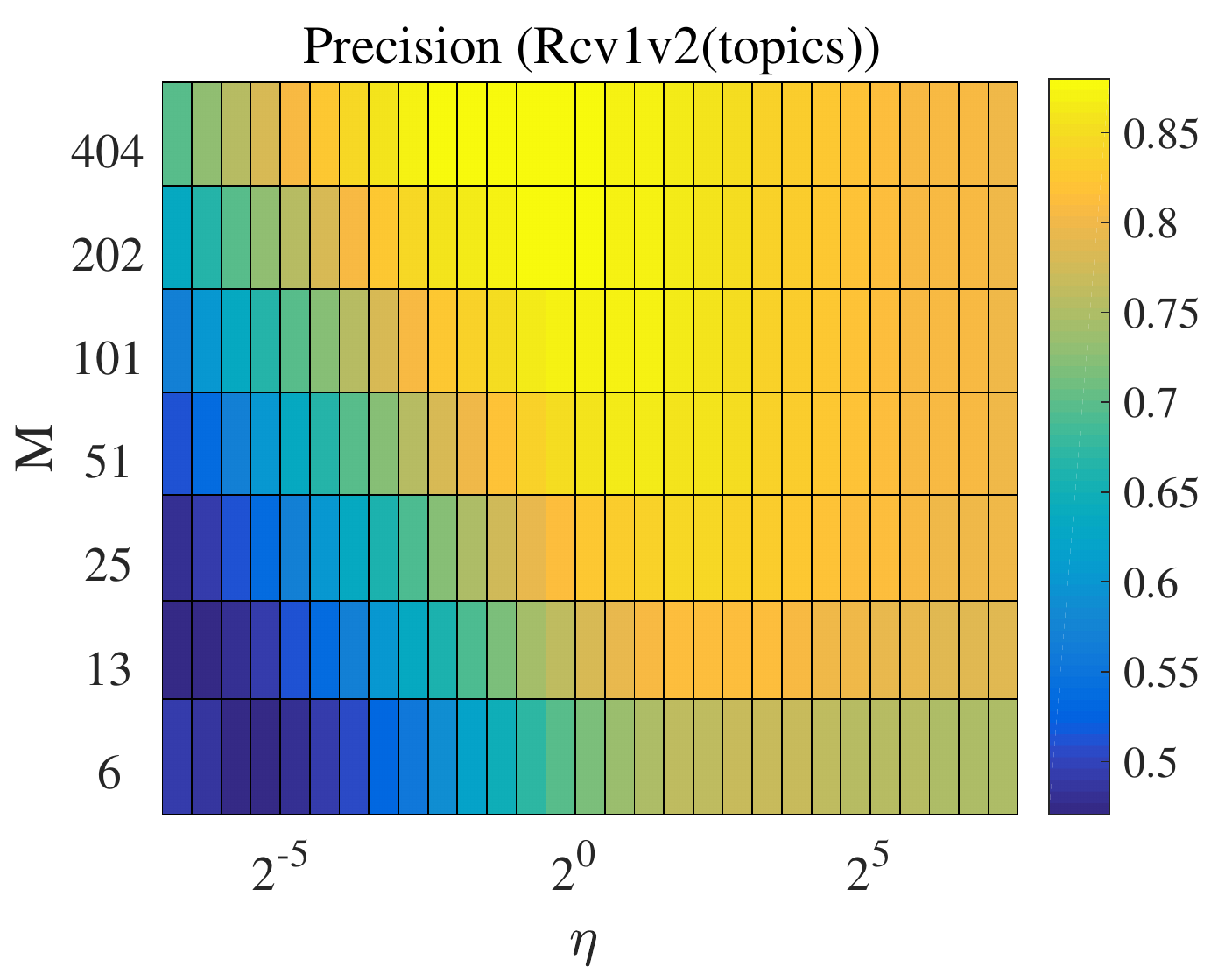}}
	\subfloat[]{\includegraphics[width=0.33 \textwidth]{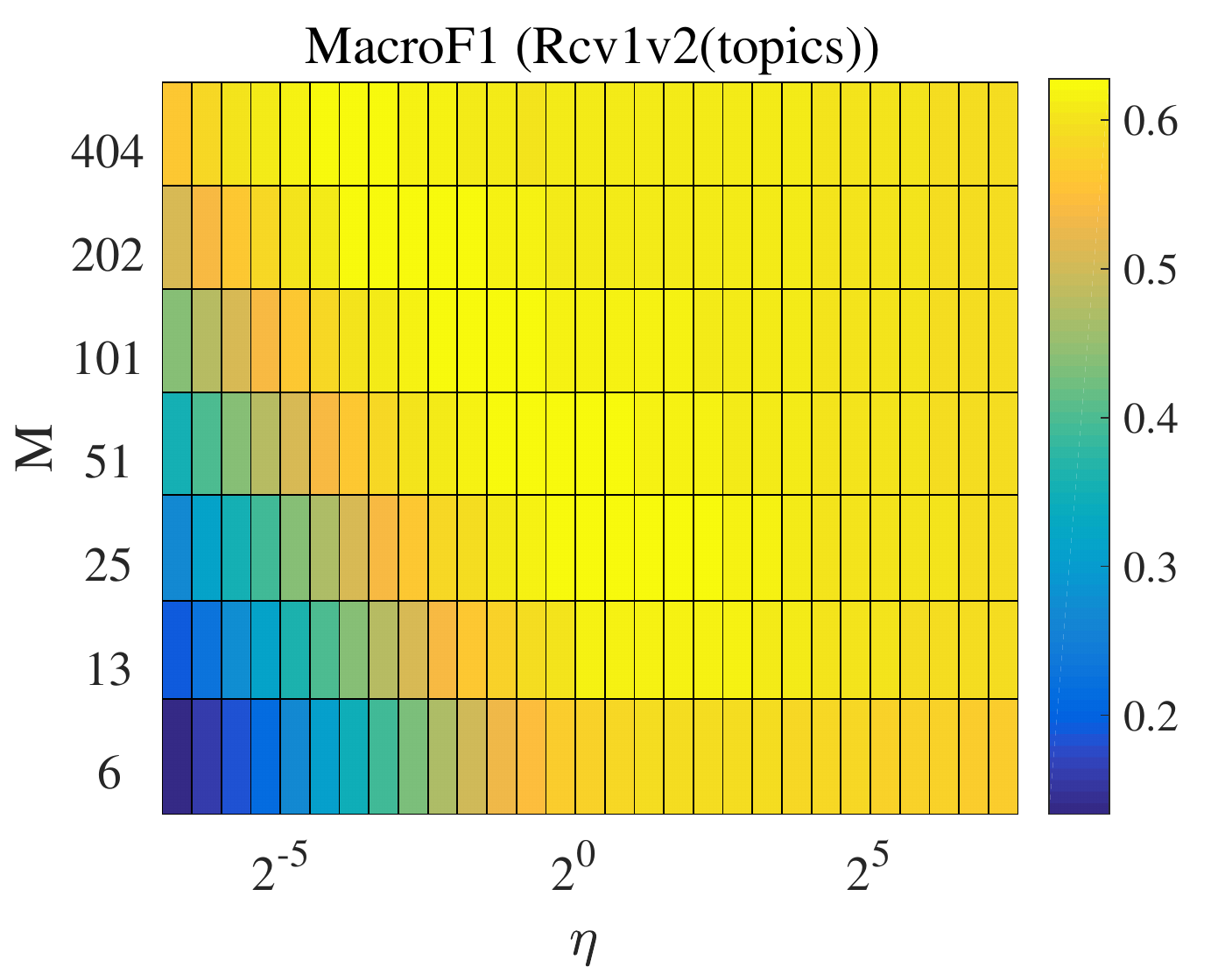}}
	\subfloat[]{\includegraphics[width=0.33 \textwidth]{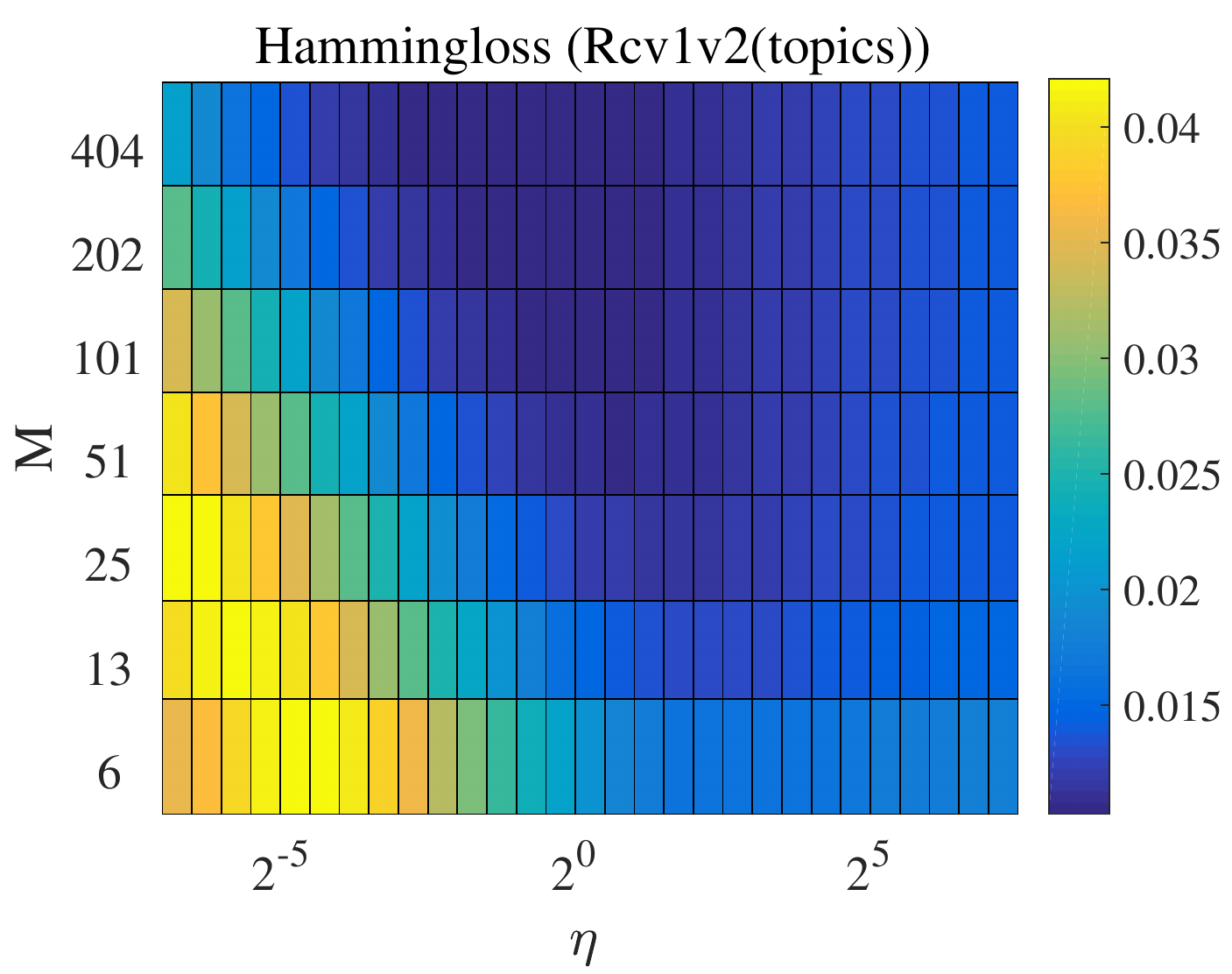}}

	\subfloat[]{\includegraphics[width=0.33 \textwidth]{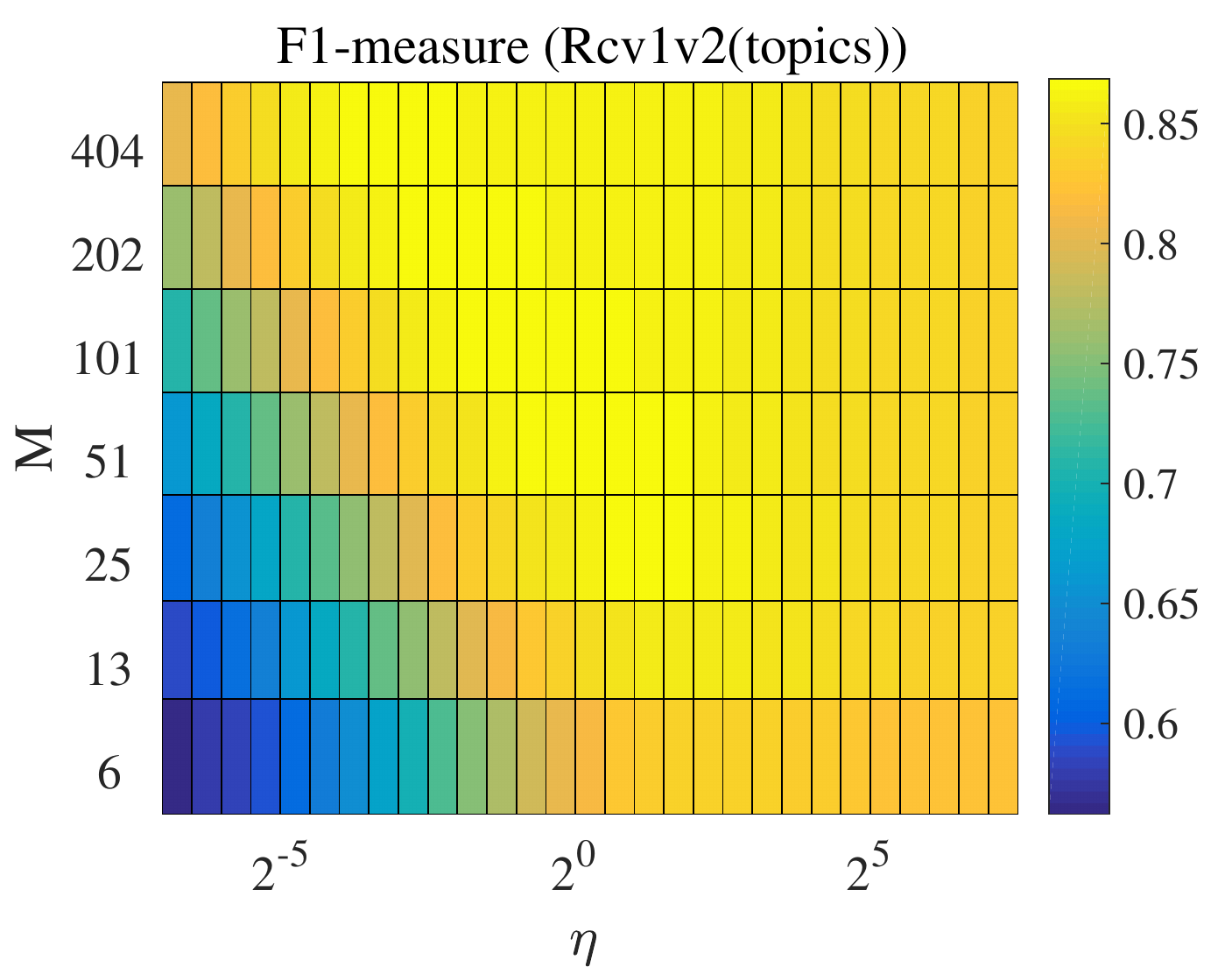}}
	\subfloat[]{\includegraphics[width=0.33 \textwidth]{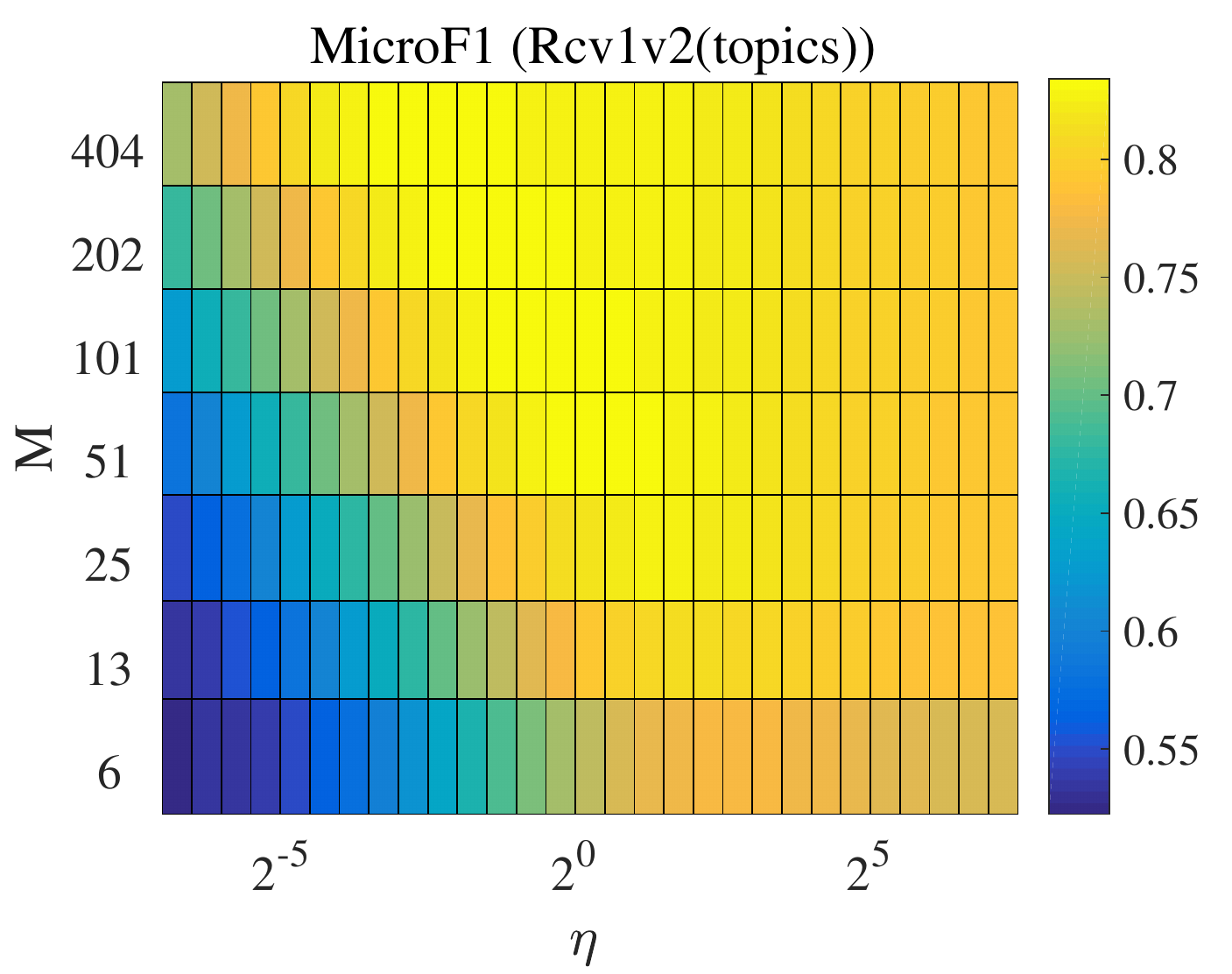}}
	\subfloat[]{\includegraphics[width=0.33 \textwidth]{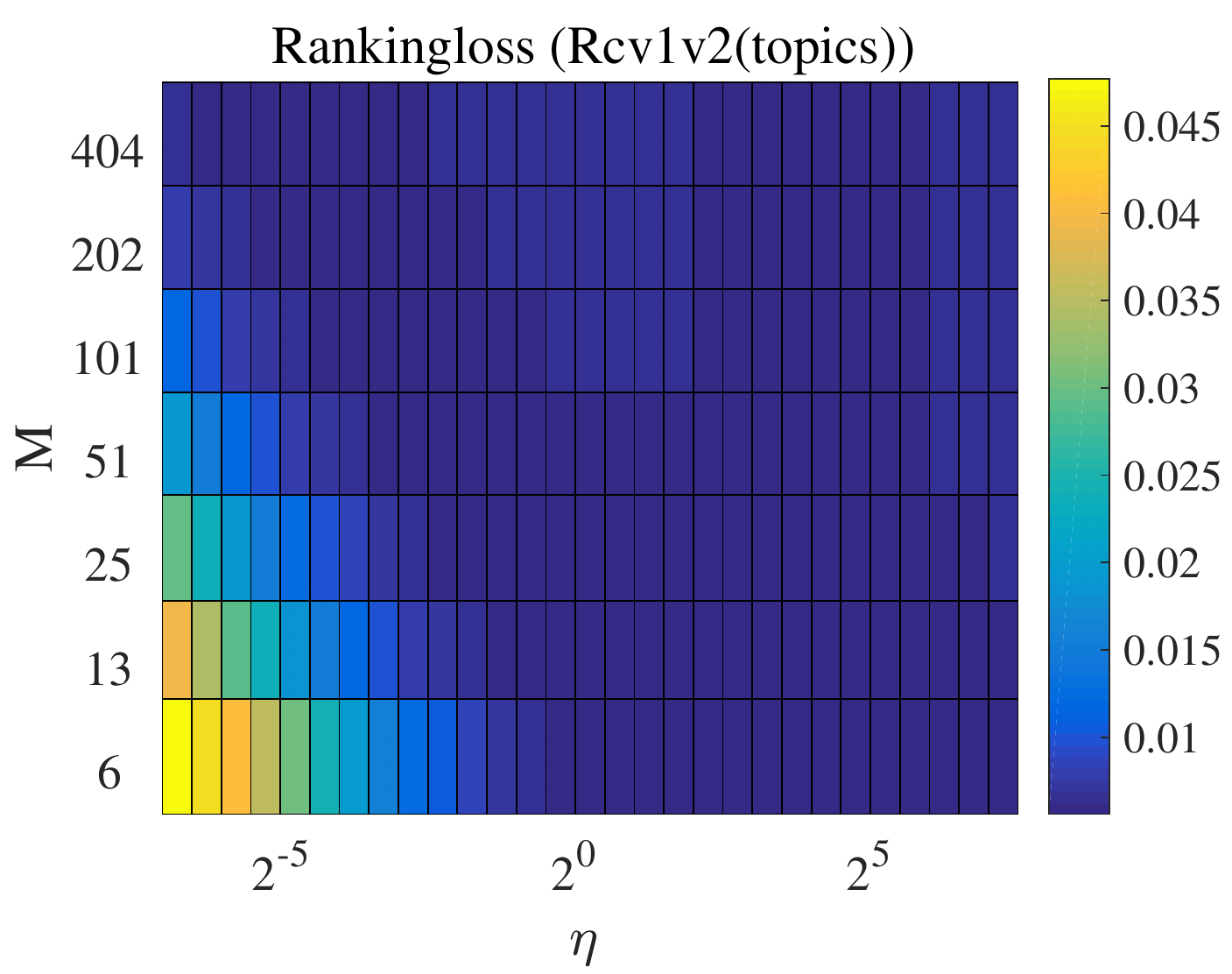}}
	\caption{Performance metrics achieved by FALT(l) using different $\eta$ and $M$.}
	\label{sensitivity_FALT}
\end{figure*}

\begin{figure*}[htpb]
	\centering
	\subfloat[]{\includegraphics[width=0.33 \textwidth]{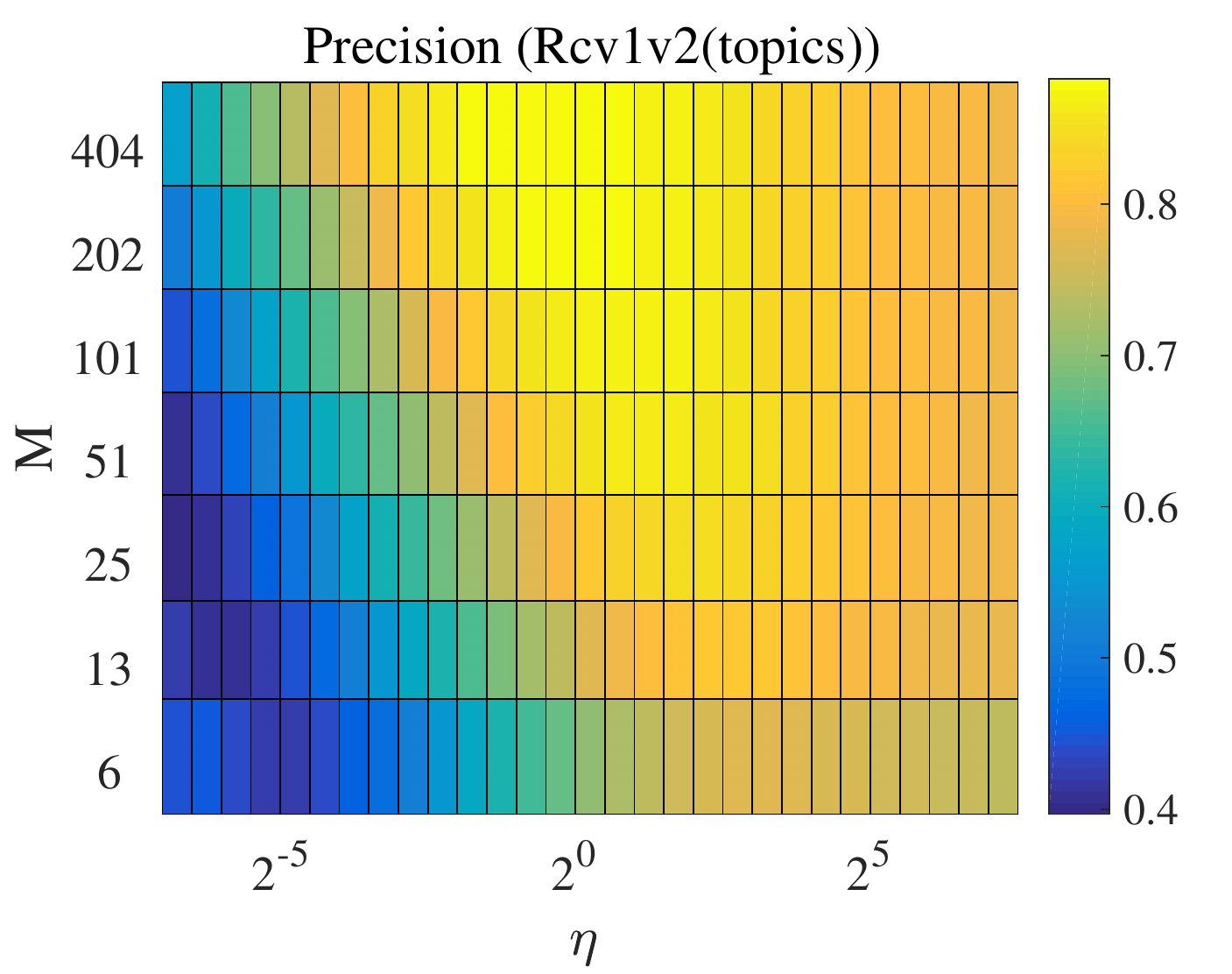}}
	\subfloat[]{\includegraphics[width=0.33 \textwidth]{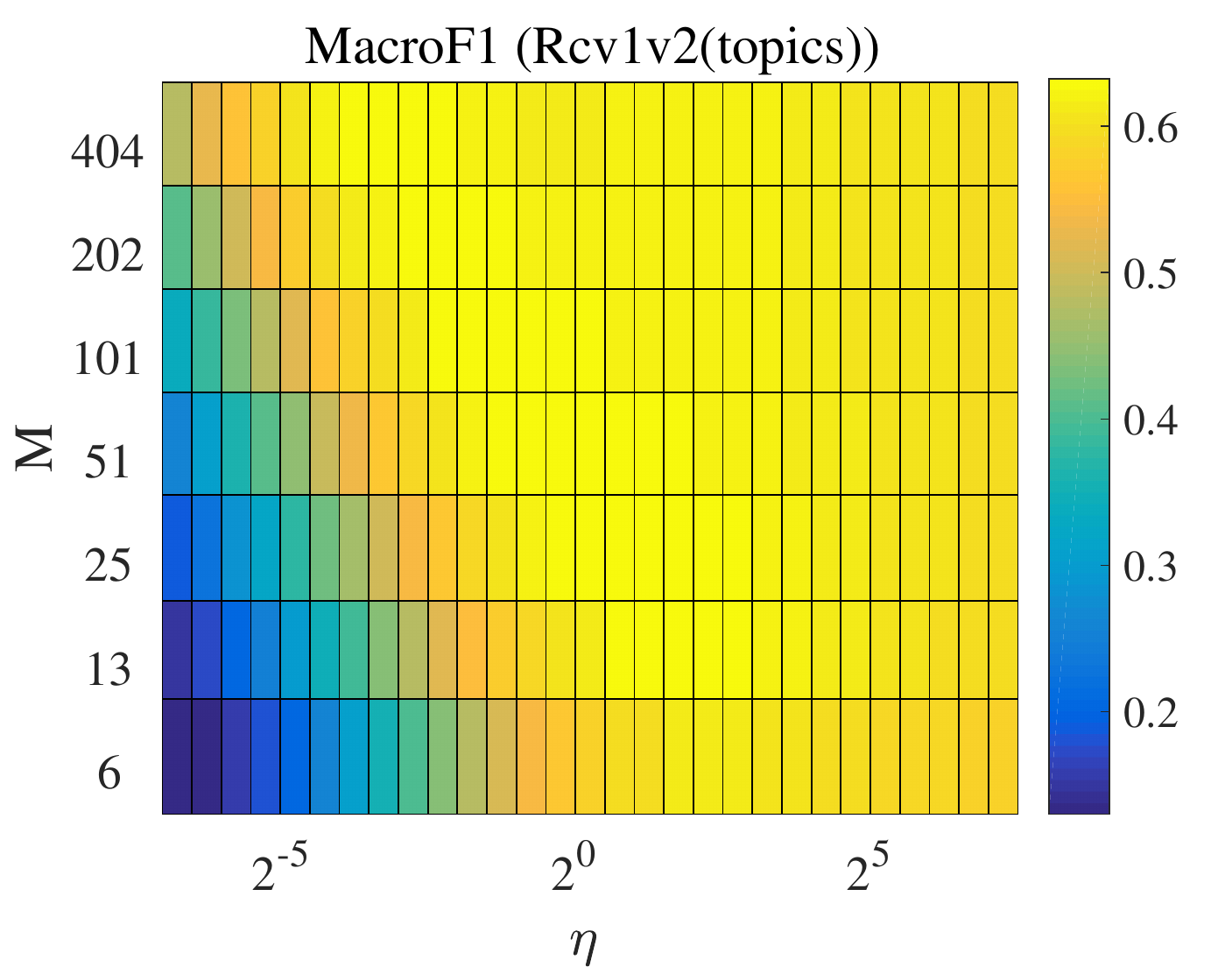}}
	\subfloat[]{\includegraphics[width=0.33 \textwidth]{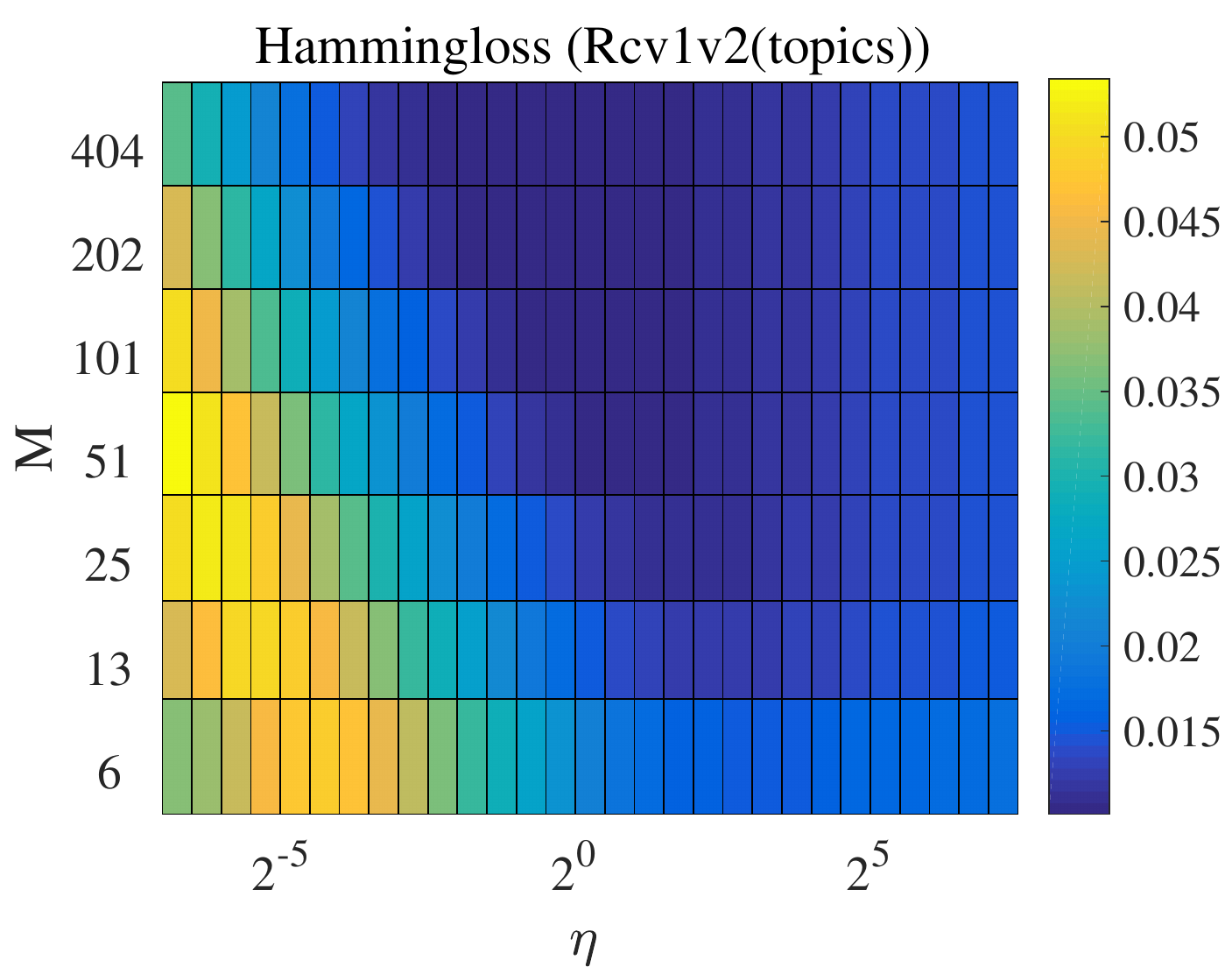}}

	\subfloat[]{\includegraphics[width=0.33 \textwidth]{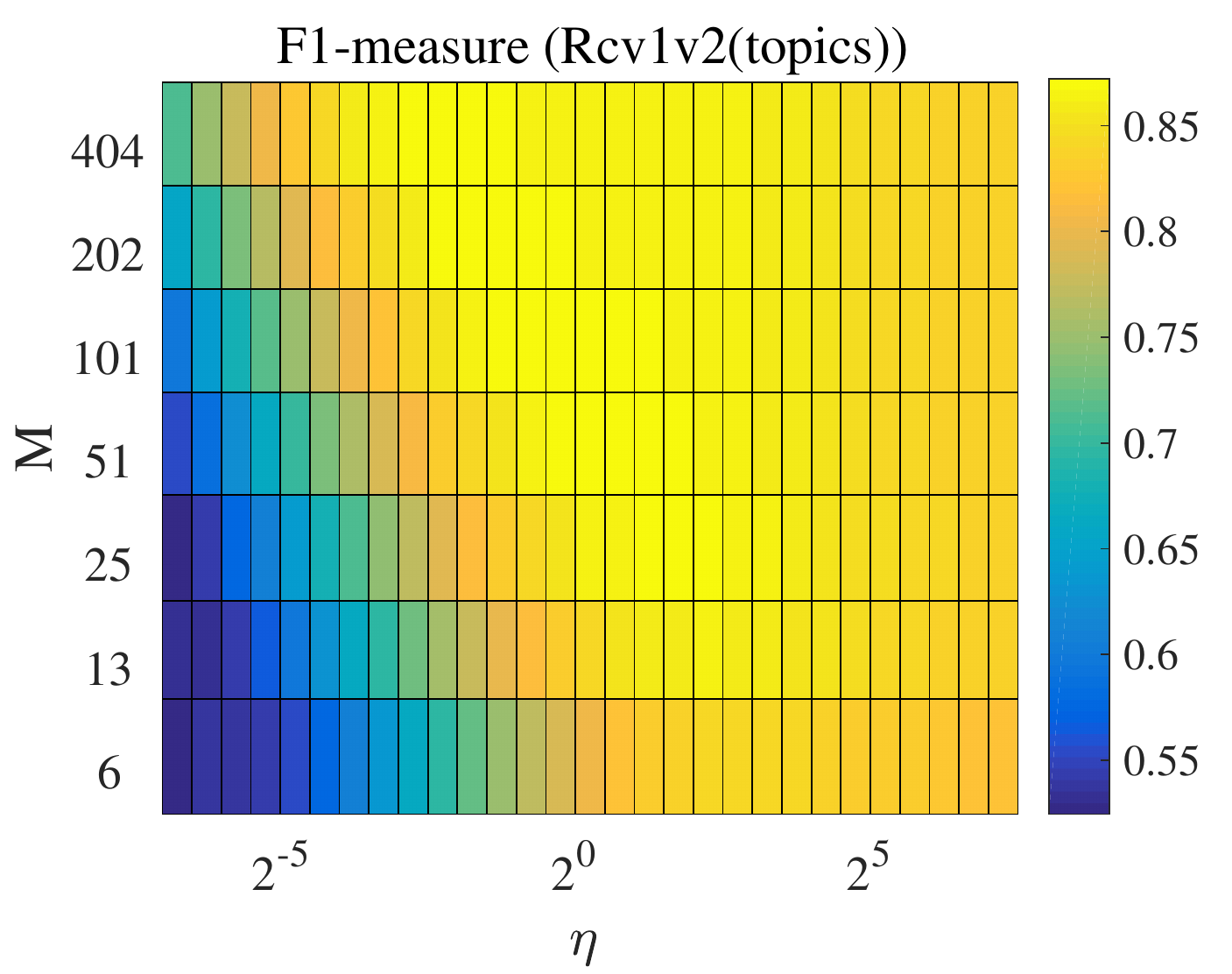}}
	\subfloat[]{\includegraphics[width=0.33 \textwidth]{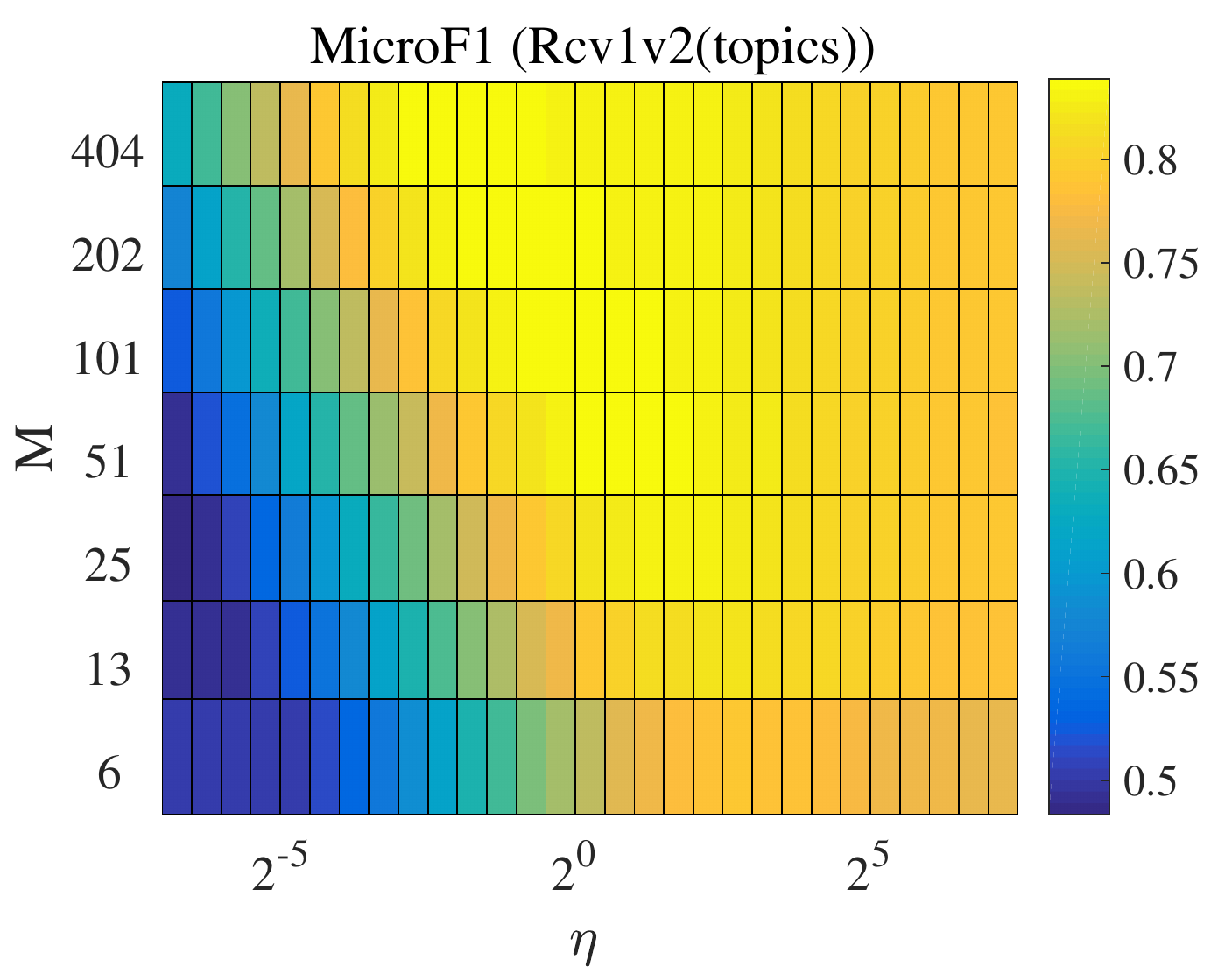}}
	\subfloat[]{\includegraphics[width=0.33 \textwidth]{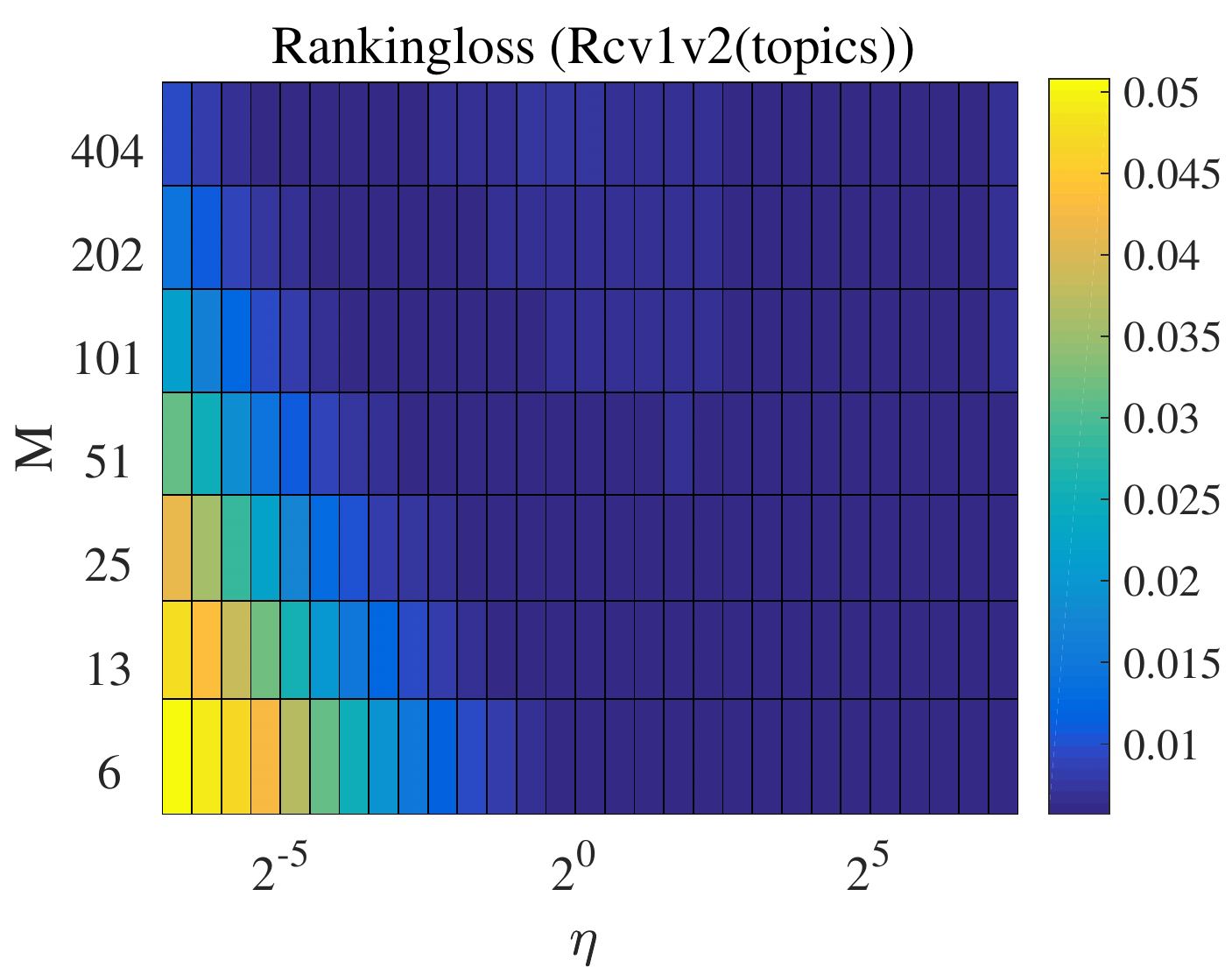}}
	
	\subfloat[]{\includegraphics[width=0.33 \textwidth]{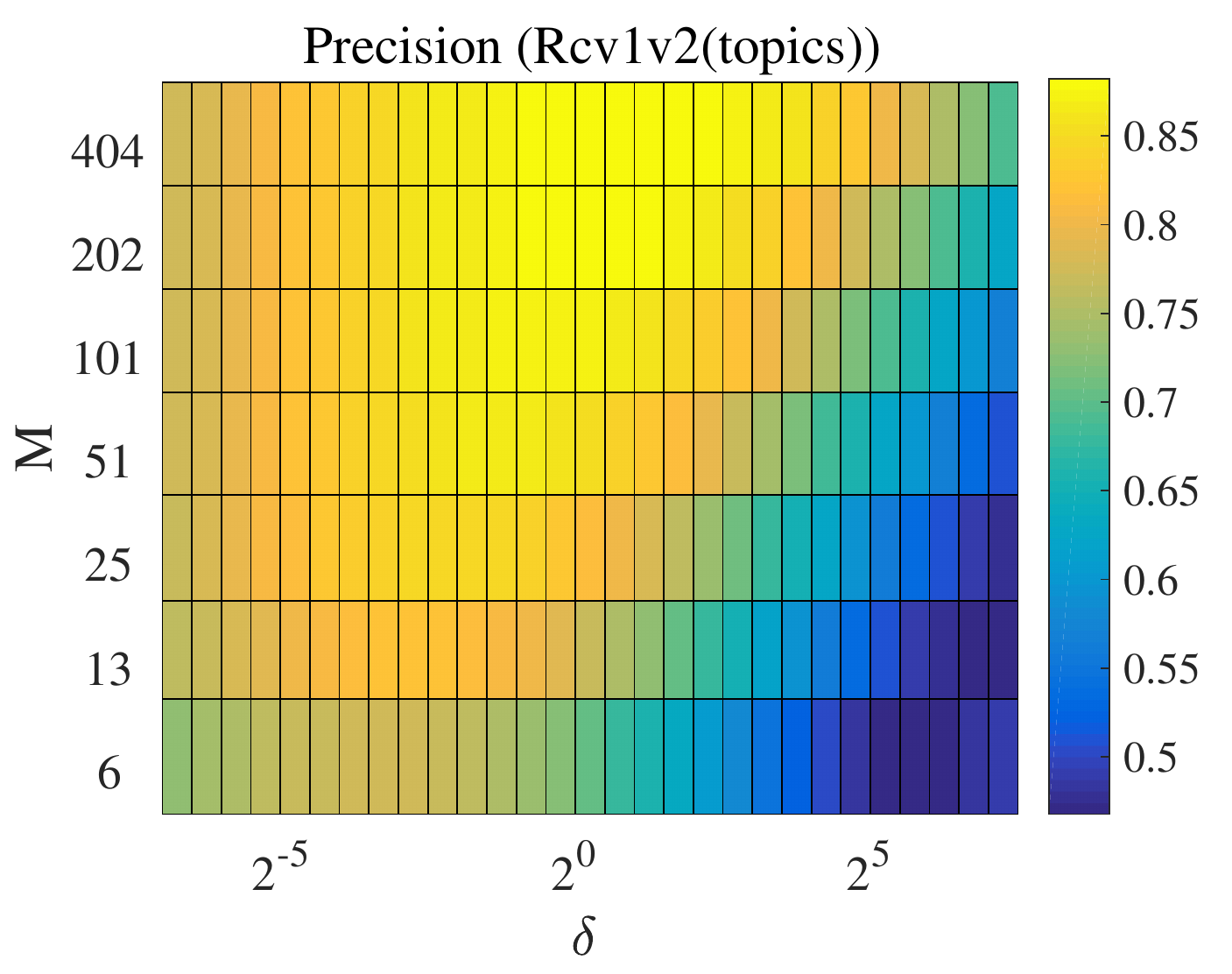}}
	\subfloat[]{\includegraphics[width=0.33 \textwidth]{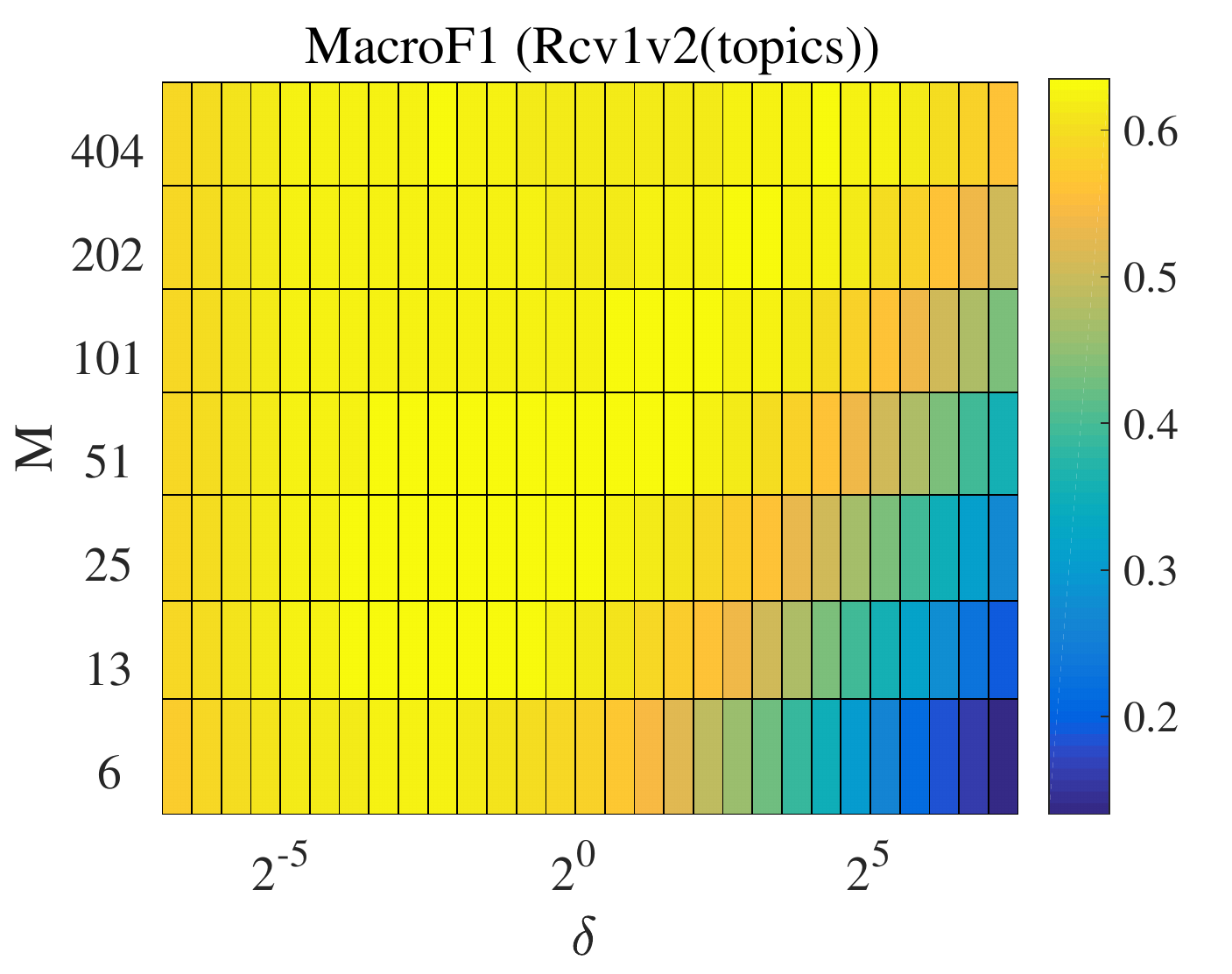}}
	\subfloat[]{\includegraphics[width=0.33 \textwidth]{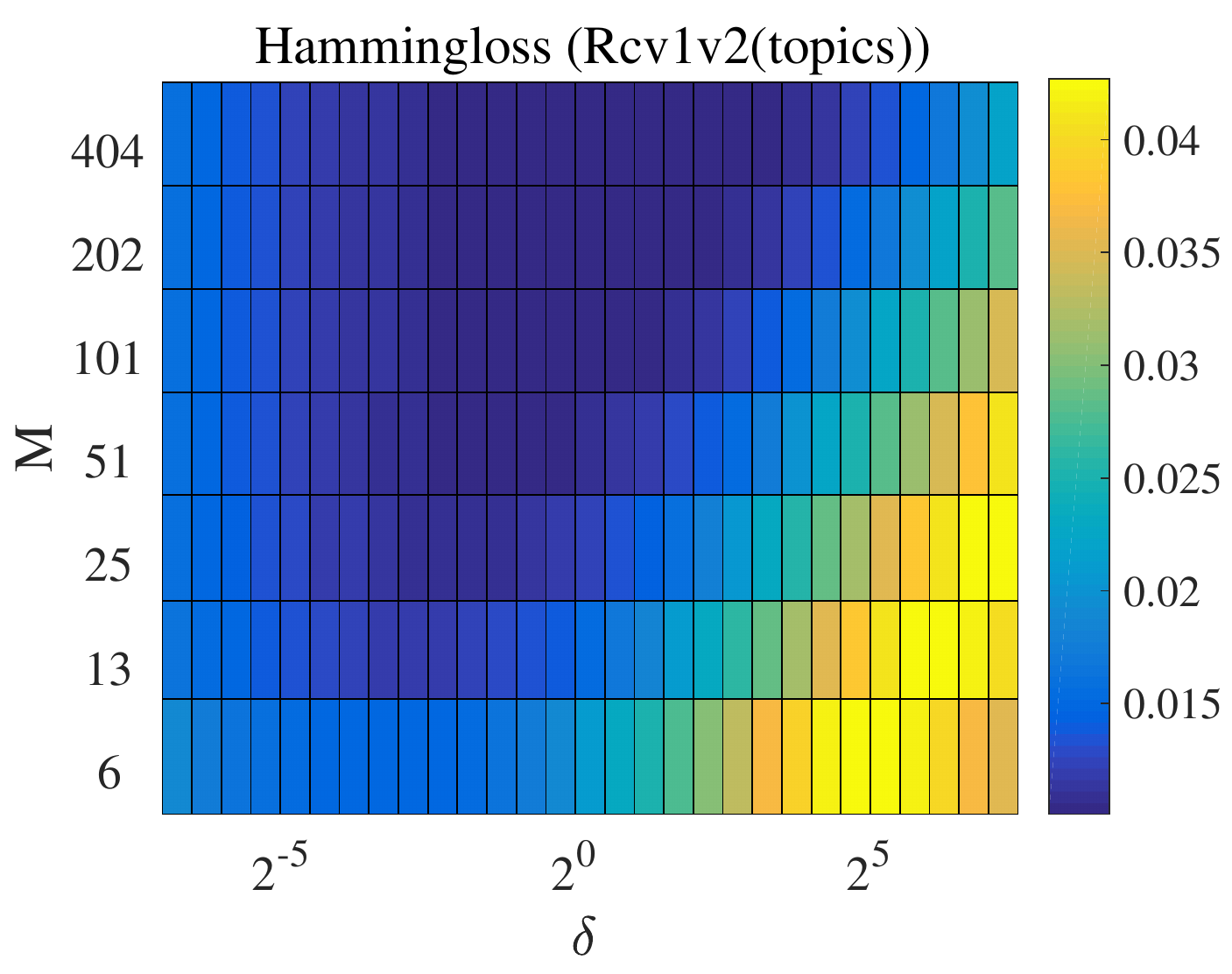}}

	\subfloat[]{\includegraphics[width=0.33 \textwidth]{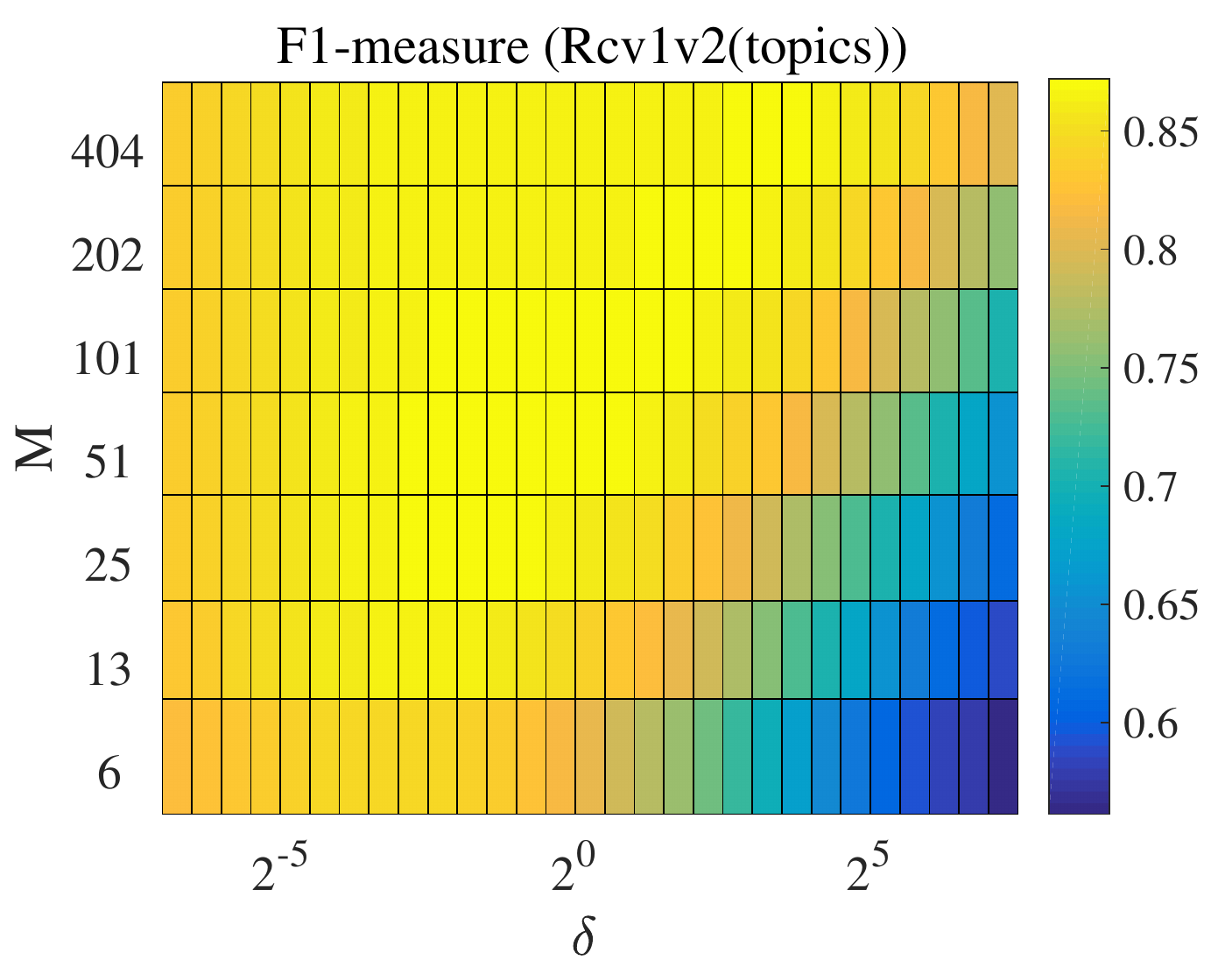}}
	\subfloat[]{\includegraphics[width=0.33 \textwidth]{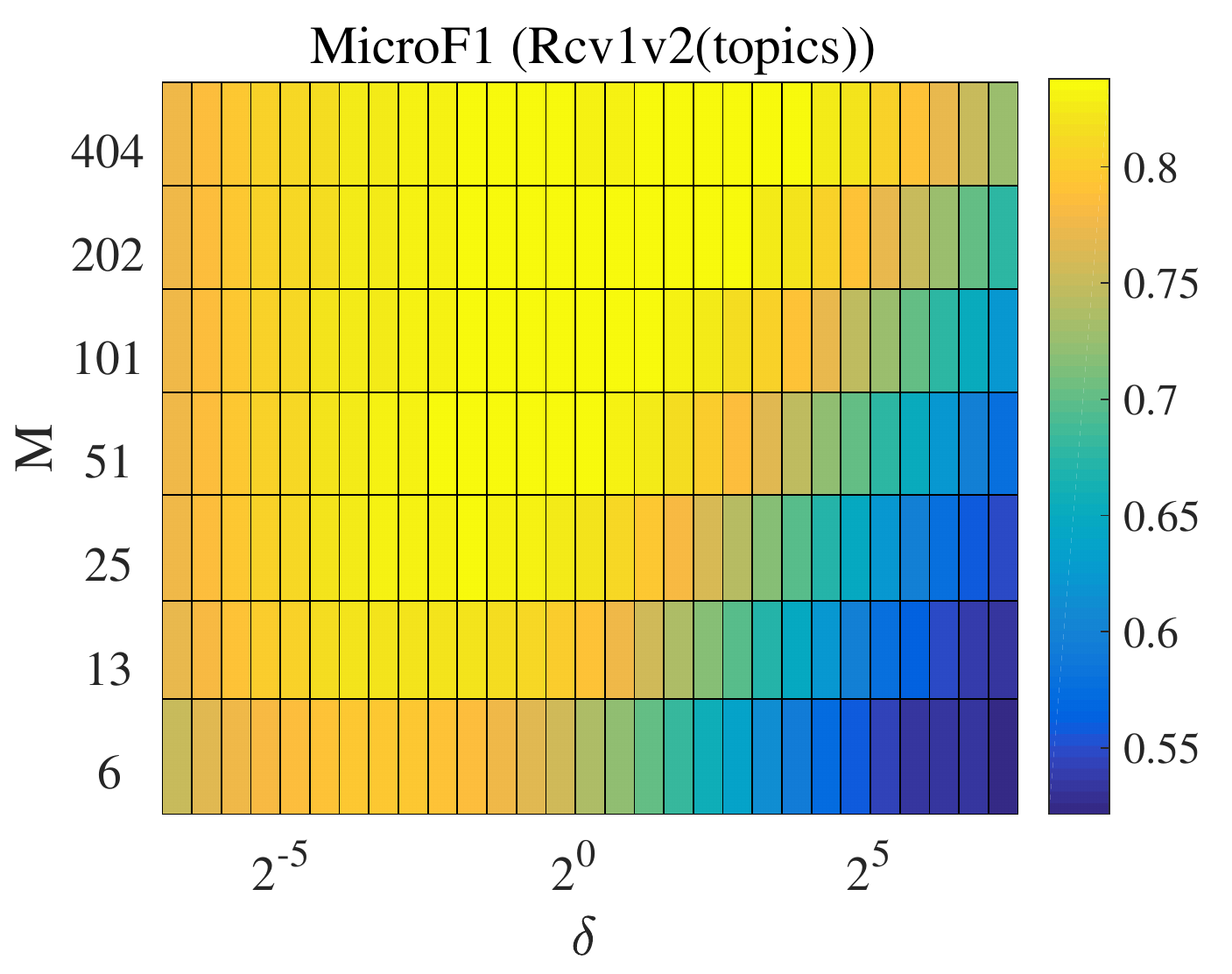}}
	\subfloat[]{\includegraphics[width=0.33 \textwidth]{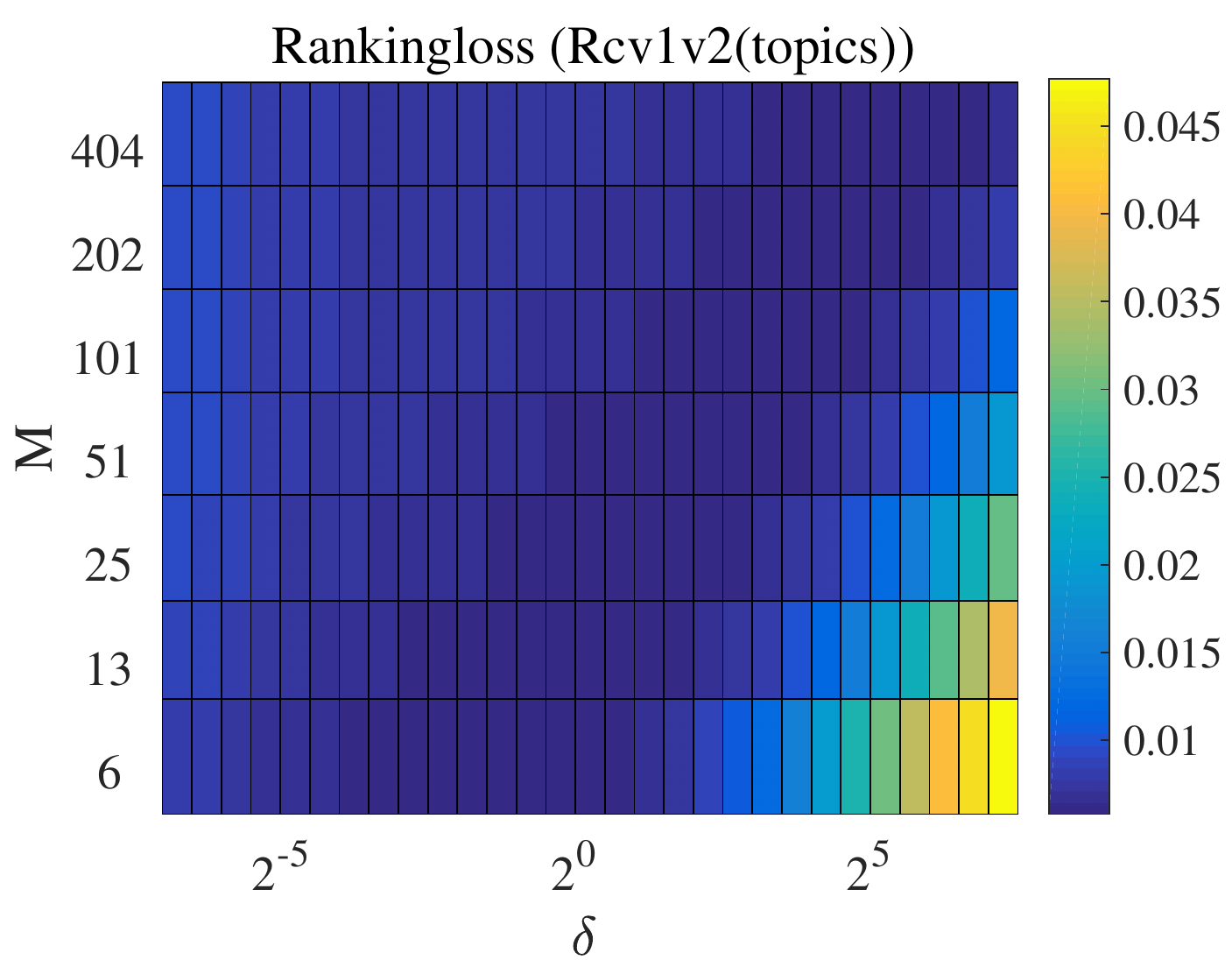}}
	\caption{Performance metrics achieved by SALT using different $\eta$ and $M$ when $\delta = 1$ (the first six subfigures) and performance metrics by SALT using different $\delta$ and $M$ when $\eta = 1$ (the last six subfigures).}
	\label{sensitivity_SALT}
\end{figure*}

Fig.~\ref{sensitivity_FALT} displays how various metrics attained by FALT(l) vary with different step-size $\eta$ and the learning times $M$ on \textsl{Rcv1v2(topics)}, where different colors represent different metric values that are obtained by ten-fold cross validation on the training set. 
Similarly, Fig.~\ref{sensitivity_SALT} also reports metric values attained by SALT on \textsl{Rcv1v2(topics)} using different $\eta$ and $M$ when $\delta = 1$ and the metric values using different $\delta$ and $M$ when $\eta= 1$. 

Due to the space constraint, we cannot display the results on the other datasets. 
However, we reach similar conclusions from all results:
\begin{enumerate}
\item 
The optimal hyperparameter regions for different metrics are possibly different. 
If these optimal regions share a common sub region, a set of good hyperparameters can be found to optimize all metrics.
Otherwise, one has to make a compromise when searching for the optimal set of hyperparameters for all metrics. 
Therefore, in our hyperparameter optimization procedure, if a set of hyperparameters achieves better results than the other set of hyperparameters in some of all performance metrics, but worse results in the other metrics, then the set of hyperparameters that gets better results in more metrics will be kept.

\item
Smaller $M$ leads to worse results in many metrics.
As $M$ increases, the hyperparameter ranges that can attain good metric values grow large.
This observation applies to FALT(l) and SALT.
Thus, in our comparative experiments, we allow to take a large value of $M$ for our algorithms.

\item
When $M$ is fixed, $\eta$ in FALT(l) and SALT should be neither too large nor too small.
This observation is consistent with their regret bounds.
In the regret bound (\ref{regret}) of FALT, when $\eta$ is too large, the term proportional to $\eta$ dominates, which leads to high regret and deteriorates the performance; conversely, when $\eta$ is too small, another term proportional to $1/\eta$ dominates, which also leads to high regret. The analysis holds for SALT. 
Thus, we suggest searching the optimal $\eta$ around $1$ for both algorithms.

\item
Lastly, when $\eta$ is fixed, $\delta$ in SALT should also be searched around 1.
This can be explained according to the regret bound (\ref{regret1}) of SALT.
On the one hand, too large $\delta$ leads to large $\frac{\delta}{2 \eta}\|\bm U_*\|_{F}^2$, which leads to high regret.
On the other hand, too small $\delta$ leads to large $(\bm H_T^{(i)})^{-1}$, which produces large $\bm w_t^{(i)}$ and further large $Q$ in (\ref{regret1}).
\end{enumerate}

\subsection{Performance comparisons}
\label{sec.testexperiments}

We run all the compared algorithms on the training datasets in table \ref{data} and 
evaluate the obtained model on the corresponding testing dataset.
The model for evaluation is obtained after a single pass through the training data.

For OSML-ELM and ELM-OMLL, the top 20\% examples in each training set is used for network initialization and
the number of hidden layer neurons is chosen from $[100, 1000]$.
The label threshold for OSML-ELM is set to 0 \cite{VenkatesanEDPW17}.
The regularization factor for ELM-OMLL is chosen from $[2^{-10}, 2^{10}]$.
The aggressiveness coefficient for PA-I-BR(l) and PA-II-BR(l) is searched from $[2^{-15}, 2^{10}]$. 
For FALT(l) and SALT, the step-size $\eta$ is chosen from $[2^{-10}, 2^{10}]$ and 
the learning times $M$ is from $[2^{-3}, 2^2]\cdot L$.
For SALT, $\delta$ is chosen from $[2^{-10}, 2^{10}]$.
For PA-I-BR(k), PA-II-BR(k) and FALT(k), an additional kernel hyperparameter $\delta^2$ is searched from $[2^{-10}, 2^{10}]$. 
Hyperparameter optimization is done by performing ten-fold cross validation on each training set, 
in which, only one pass over the training splits is allowed. Each algorithm is run and evaluated $20$ times on each dataset, each time with a different permutation of examples in the training set. 
Performance metrics achieved by each algorithm on each testing set are averaged over 20 runs, and reported in Table \ref{result1},
where the best result in each metric and its comparable ones are displayed in bold, 
according to paired t-tests at 95\% confidence level.
Moreover, the average training and testing time in seconds for each algorithm is reported in Table \ref{result2}.

\begin{table*}[!htp]
	\renewcommand{\arraystretch}{0.7}
	\caption{Testing performance metrics [\%] and standard derivation (in brackets) achieved by all compared algorithms}
	\label{result1}
	\centering \tabcolsep = 0.1pt
	\begin{threeparttable}
	\footnotesize
	\begin{tabular}{lcccccccc}
	\hline
Dataset & Algorithm  & Psn  & Rcal  & F1   & MacroF1   & MicroF1   & Hl  & Rl \\
	\hline
	\multirow{2}{*}{}
	 &OSML-ELM &57.22 (0.06) &56.93 (0.04) &57.08 (0.05)  &0.34 (0.02) &2.91 (0.18)  &0.22 (0.00) &4.13 (0.04)  \\
	 &ELM-OMLL &63.72 (0.16) &61.77 (0.15) &62.73 (0.15)  &8.24 (0.09) &34.91 (0.22)  &0.21 (0.00)  &4.13 (0.04)  \\
	 {Rcv1v2}
	 &PA-I-BR(l) &69.21 (0.25) &67.47 (0.23) &68.33 (0.24)   &20.13 (0.33) &42.28 (0.72) &\textbf{0.18 (0.00)} &2.54 (0.05)  \\
	 {(industries)}
	 &PA-II-BR(l) &69.21 (0.25) &67.46 (0.23) &68.33 (0.24)  &20.12 (0.33) &42.28 (0.72) &\textbf{0.18 (0.00)} &2.53 (0.05) \\
	 &FALT(l) &71.10 (0.19) &70.61 (0.40) &70.85 (0.27) &\textbf{23.32 (0.57)} &47.74 (0.62) &0.19 (0.00) &\textbf{1.97 (0.02)} \\
	 &SALT&\textbf{71.22 (0.13)} &\textbf{70.68 (0.29)} &\textbf{70.95 (0.19)} &\textbf{23.38 (0.47)} &\textbf{48.08 (0.55)} &0.19 (0.00) &2.00 (0.02) \\
	\hline
	\multirow{2}{*}{}
      &OSML-ELM &50.07 (0.33) &46.87 (0.31) &48.42 (0.32)  &9.25 (0.09) &54.93 (0.31) &0.37 (0.00) &1.23 (0.03)  \\
      &ELM-OMLL &70.86 (0.22) &84.68 (0.15) &77.16 (0.16)  &28.65 (0.11) &70.74 (0.16) &0.37 (0.00) &1.23 (0.03)  \\
      {Rcv1v2}
      &PA-I-BR(l) &89.25 (0.20) &86.50 (0.48) &87.85 (0.33)   &47.06 (0.32) &85.02 (0.13) &\textbf{0.16 (0.00)} &0.35 (0.02)  \\
      {(regions)}
      &PA-II-BR(l) &89.24 (0.20) &86.49 (0.48) &87.85 (0.32) &47.05 (0.32) &85.02 (0.13)  &\textbf{0.16 (0.00)} &0.35 (0.02)  \\
      &FALT(l) &89.94 (0.33) &\textbf{90.11 (0.63)} &\textbf{90.02 (0.22)} &\textbf{49.84 (0.94)} &85.53 (0.29) &\textbf{0.16 (0.01)} &\textbf{0.23 (0.01)}  \\
      &SALT &\textbf{89.99 (0.30)} &90.00 (0.61) &\textbf{89.99 (0.24)} &49.70 (0.86) &\textbf{85.58 (0.24)} &\textbf{0.16 (0.00)} &0.24 (0.01) \\
	\hline
	\multirow{2}{*}{}
	&OSML-ELM &83.57 (0.22) &54.95 (0.22) &66.31 (0.20)  &19.22 (0.15) &64.06 (0.20) &1.79 (0.01) &2.94 (0.03)  \\
	&ELM-OMLL &76.52 (0.19) &71.88 (0.18) &74.13 (0.12)  &30.91 (0.17) &70.80 (0.13) &1.77 (0.01)   &2.94 (0.03)  \\
	Rcv1v2
	&PA-I-BR(l) &88.23 (0.30) &78.95 (0.47) &83.33 (0.18)  &{53.77 (0.27)} &79.83 (0.12) &1.20 (0.01) &1.30 (0.03)   \\
	(topics)
	&PA-II-BR(l) &\textbf{88.72 (0.27)} &78.72 (0.43) &83.42 (0.16) &53.13 (0.30) &79.94 (0.13) &\textbf{1.19 (0.01)} &1.15 (0.02) \\
       	&FALT(l) &84.69 (1.03) &83.63 (0.98) &84.14 (0.11) &\textbf{56.56 (0.58)} &79.73 (0.27) &1.29 (0.04) &\textbf{0.79 (0.01)} \\
	&SALT  &84.86 (0.69) &\textbf{84.10 (0.59)} &\textbf{84.47 (0.09)} &\textbf{56.69 (0.54)} &\textbf{80.29 (0.18)} &1.26 (0.02) &0.86 (0.01)  \\
	\hline
	\multirow{5}{*}{Bibtex}
	&OSML-ELM &34.92 (0.34) &20.21 (0.22) &25.60 (0.26)  &7.05 (0.20) &26.89 (0.27) &\textbf{1.31 (0.00)}  &11.73 (0.23)   \\
	&ELM-OMLL &\textbf{44.34 (0.44)} &{43.30 (0.27)} &43.81 (0.32)  &28.23 (0.45) &\textbf{43.15 (0.32)} &1.59 (0.01) &11.79 (0.24)  \\
	&PA-I-BR(l) &44.53 (1.45) &37.36 (1.31) &40.61 (0.97)  &26.73 (0.86) &40.05 (0.77) &1.53 (0.06) &9.26 (0.26)  \\
	&PA-II-BR(l) &\textbf{44.71 (1.30)} &37.22 (1.32) &40.60 (0.95)  &26.71 (0.92) &40.18 (0.83) &1.51 (0.05) &9.30 (0.27)  \\
	&FALT(l) &\textbf{45.27 (2.34)} &43.69 (4.20) &44.25 (1.21) &30.13 (2.23) &41.78 (0.97) &1.67 (0.22) &\textbf{6.47 (0.19)} \\
	&SALT  &\textbf{45.42 (2.23)} &\textbf{47.57 (2.87)}  &\textbf{46.35 (0.78)} &\textbf{32.26 (1.16)} &42.54 (0.92) &1.78 (0.18) &6.61 (0.15)  \\
	\hline
	 \multirow{5}{*}{Birds}
	 &OSML-ELM &64.22 (1.33) &59.20 (1.15) &61.61 (1.17)  &23.69 (2.09) &38.96 (2.00) &\textbf{4.46 (0.12)}  &9.53 (0.61)   \\
	 &ELM-OMLL &64.24 (0.92) &64.66 (0.84) &64.45 (0.84) &33.51 (1.40) &44.05 (0.91) &5.81 (0.13) &10.16 (0.41)  \\
	 &PA-I-BR(k) &64.60 (0.00) &64.34 (0.00) &64.47 (0.00)  &36.95 (0.00) &46.18 (0.00) &5.17 (0.00)  &10.42 (0.32)  \\
	 &PA-II-BR(k) &64.60 (0.00) &64.34 (0.00) &64.47 (0.00) &37.04 (0.02) &46.33 (0.02) &5.13 (0.00)  &10.81 (0.34) \\
	 &FALT(k) &\textbf{65.19 (0.00)} &\textbf{71.42 (0.00)} &\textbf{68.16 (0.00)} &\textbf{41.43 (0.02)} &\textbf{50.01 (0.03)} &5.70 (0.01) &\textbf{9.36 (0.09)} \\
	 \hline
	 \multirow{5}{*}{Scene}
	  &OSML-ELM &61.27 (0.93) &61.57 (0.94) &61.42 (0.93)   &68.65 (0.79) &68.13 (0.76) &10.20 (0.22) &9.96 (0.34) \\
	  &ELM-OMLL &64.79 (0.87) &71.94 (0.93) &68.18 (0.85)  &69.65 (0.72) &68.57 (0.71) &11.72 (0.28) &10.69 (0.41)   \\
	  &PA-I-BR(k) &70.57 (1.75) &70.44 (2.43) &70.50 (2.05) &74.16 (1.57) &73.96 (1.43) &\textbf{8.77 (0.46)} &7.39 (0.45)  \\
	  &PA-II-BR(k)&70.43 (1.73) &70.42 (2.43) &70.43 (2.04) &74.07 (1.52) &73.83 (1.41) &8.83 (0.48) &7.47 (0.50)  \\
	  &FALT(k) &\textbf{73.12 (0.92)} &\textbf{75.25 (1.45)} &\textbf{74.17 (1.07)} &\textbf{75.83 (0.68)} &\textbf{75.24 (0.77)} &\textbf{8.77 (0.35)} &\textbf{7.18 (0.19)} \\	
	 \hline
	 \multirow{5}{*}{Emotions}
	  &OSML-ELM &66.00 (1.70)  &60.64 (1.39)  &63.20 (1.41)  &64.29 (1.27) &65.66 (1.14) &20.91 (0.64) &17.57 (0.90) \\
	  &ELM-OMLL &67.31 (1.03)  &72.85 (1.00) &69.97 (0.95)  &69.48 (0.78) &70.12 (0.76) &20.57 (0.48) &16.38 (0.42)  \\
	  &PA-I-BR(k) &67.56 (2.19) &64.74 (2.54) &66.11 (2.14)  &67.26 (1.94) &68.78 (1.52)  &\textbf{19.59 (0.75)} &\textbf{15.53 (0.91)}  \\
	  &PA-II-BR(k) &66.91 (1.97) &64.25 (1.93) &65.55 (1.77) &66.86 (1.68) &68.47 (1.26) &\textbf{19.73 (0.63)} &\textbf{15.63 (0.86)}  \\
         &FALT(k)  &\textbf{69.64 (0.67)} &\textbf{73.52 (0.96)} &\textbf{71.52 (0.53)} &\textbf{70.19 (0.49)} &\textbf{71.30 (0.39)} &\textbf{19.50 (0.27)} &\textbf{15.54 (0.16)} \\
	 \hline
	 \multirow{5}{*}{Yeast}
	 &OSML-ELM &69.20 (0.48) &57.74 (0.53) &62.95 (0.47)  &35.39 (0.47) &62.85 (0.48) &20.44 (0.23) &18.34 (0.16)  \\
	 &ELM-OMLL &68.58 (0.36) &62.37 (0.38) &65.33 (0.32) &39.86 (0.36) &65.10 (0.27) &20.01 (0.14) &17.76 (0.14)   \\
	 &PA-I-BR(k) &\textbf{71.59 (0.71)} &58.97 (1.62) &64.65 (0.89)  &40.30 (0.61) &64.85 (0.62) &\textbf{19.22 (0.17)} &16.24 (0.16)  \\
	 &PA-II-BR(k)&\textbf{71.73 (0.86)} &58.91 (1.78) &64.67 (0.93) &39.66 (0.76)  &64.84 (0.67) &\textbf{19.20 (0.20)} &16.29 (0.17)  \\
	 &FALT(k)  &68.63 (0.80) &\textbf{67.86 (1.25)} &\textbf{68.23 (0.29)} &\textbf{42.60 (0.80)} &\textbf{67.57 (0.23)} &19.48 (0.25) &\textbf{16.03 (0.20)}  \\
	 \hline
	 \multirow{5}{*}{Mediamill}
	 &OSML-ELM &\textbf{75.89 (0.19)} &48.14 (0.11) &58.91 (0.13)  &7.27 (0.09) &56.49 (0.11) &\textbf{2.98 (0.01)}  &6.18 (0.08) \\
	 &ELM-OMLL &69.20 (0.06) &53.28 (0.06) &60.20 (0.05) &6.29 (0.02) &57.94 (0.04) &3.16 (0.00) &5.14 (0.01)  \\
	 &PA-I-BR(k) &72.58 (0.13) &52.87 (0.13) &61.17 (0.07) &23.71 (0.21) &58.78 (0.09) &3.01 (0.00) &5.80 (0.05)  \\
	 &PA-II-BR(k)&72.50 (0.14) &52.89 (0.13) &61.16 (0.07) &\textbf{23.82 (0.20)} &58.77 (0.09) &3.02 (0.00) &5.83 (0.05) \\
	 &FALT(k)     &70.63 (0.12) &\textbf{54.78 (0.09)} &\textbf{61.70 (0.05)}  &\textbf{23.90 (0.17)} &\textbf{59.43 (0.05)} &3.04 (0.00) &\textbf{4.76 (0.01)}  \\ 
	\hline
	\end{tabular}
	\end{threeparttable}	
\end{table*}

\begin{table*}[!htp]
	\renewcommand{\arraystretch}{0.7}
	\caption{Training and testing time spent by all algorithms}
	\label{result2}
	\centering \tabcolsep = 1pt
	\footnotesize
	\begin{tabular}{lcccccccccc}
	\hline
        \multirow{2}{*}{Algorithm}  &\multicolumn{2}{c}{Rcv1v2(industries)}  &\multicolumn{2}{c}{Rcv1v2(regions)} &\multicolumn{2}{c}{Rcv1v2(topics)}  &\multicolumn{2}{c}{Bibtex} &\multicolumn{2}{c}{}\\
         \cline{2-11}
          &Train[s]  &Test[s]  &Train[s]   &Test[s]  &Train[s]   &Test[s]  &Train[s]  &Test[s]  &&\\
	\hline
	 OSML-ELM  &4903.34 &656.04  &4832.96 &643.34 &4861.00 &640.60 &94.43 &0.13 &&\\
	 ELM-OMLL  &4750.08 &641.78   &4866.87 &642.53 &4934.48  &644.90 &100.42 &0.14 &&\\
	 PA-I-BR(l)  &3.84 &110.97  &2.54 &88.53  &0.94 &51.08  &0.09 &0.11 &&\\
	 PA-II-BR(l)  &3.85 &109.80 &2.70 &90.20  &0.92 &47.31  &0.10 &0.11 &&\\
	 FALT(l)    &6.36 &105.75 &3.80 &86.15  &1.68 &48.22      &0.32 &0.11 &&\\
	 SALT      &14.17 &113.58 &7.80 &88.93 &3.99 &48.94      &0.71  &0.12 &&\\
	\hline
        \multirow{2}{*}{Algorithm}  &\multicolumn{2}{c}{Birds}  &\multicolumn{2}{c}{Scene} &\multicolumn{2}{c}{Emotions}  &\multicolumn{2}{c}{Yeast} &\multicolumn{2}{c}{Mediamill}\\
         \cline{2-11}
        &Train[s]  &Test[s]  &Train[s]   &Test[s]  &Train[s]   &Test[s]  &Train[s]  &Test[s] &Train[s]  &Test[s]\\
        \hline
	 OSML-ELM   &2.84 &0.01  &13.14 &0.02 &0.67 &0.00  &0.60 &0.01  &284.13  &0.30 \\
	 ELM-OMLL   &1.73 &0.01  &2.24 &0.01  &0.96 &0.00  &15.18 &0.02  &306.78  &0.30 \\
	 PA-I-BR(k)   &0.01 &0.01  &0.02 &0.04  &0.01 &0.00   &0.03 &0.03  &131.91 &10.27 \\
	 PA-II-BR(k)  &0.01 &0.01  &0.02 &0.04  &0.01  &0.00  &0.03 &0.03  &139.04 &10.05\\
	 FALT(k)       &0.05 &0.01  &0.03 &0.04  &0.01 &0.00   &0.07 &0.04  &160.73 &10.16 \\ 
	\hline
	\end{tabular}
\end{table*}

From Table \ref{result1}, on the first four datasets, we observe that FALT(l) and SALT outperform the other algorithms in terms of most performance metrics, which shows that they can well address the problem of online label thresholding.
We also observe that SALT is slightly better than or at least comparable to FALT(l) in most metrics.
Two ELM-based methods perform particularly poorly on the three \textsl{Rcv1v2} datasets, which is due to the conflict that the datasets are highly feature-sparse but the sample complexity for reaching their peak performance is high.
On the remaining five datasets, we further observe that FALT(k) defeats the other algorithms in most metrics, and it always achieves the best results in F1-measure, MacroF1, MicroF1 and Ranking loss.
Especially, the differences in some metrics between FALT(k) and the second best algorithm are above 3\%.
The promising results demonstrate the great benefit by equipping FALT with the RBF kernel.

According to Table \ref{result2}, in terms of training time, two BR-based methods run the fastest, then it comes our proposed algorithms, FALT(l), SALT or FALT(k), and two ELM-based methods run the slowest. FALT(l) runs faster than SALT.
In terms of testing time, our algorithms and two BR-based methods take similar time. They predict faster than two ELM-based methods in the linear version, but slower in the kernelized version when the number of support vectors they kept is larger,
such as on \textsl{Mediamill}.

In summary, in terms of comprehensive performance, our proposed algorithms outperform the other ones,
since they achieve the most competitive results in terms of most metrics, meanwhile spend relatively less time.

\section{Conclusion}
\label{sec.conclusion}

In this paper, a novel algorithmic framework of adaptive label thresholding is proposed for handling the online label thresholding problem. Based on this framework, we proposed two algorithms, namely, FALT and SALT.
Both algorithms exploit a novel multi-label classification loss function to measure to what an extent the multi-label classifier can separate relevant labels from irrelevant ones for an incoming instance. 
FALT updates its multi-label classifier with online gradient descent method, while SALT with adaptive mirror descend method.
Both FALT and SALT are proved to enjoy a sub-linear regret.
FALT has been generalized to the nonlinear multi-label prediction tasks using the kernel trick.
We have shown the superiority of the linear FALT and SALT and the nonlinear FALT with RBF kernel, in terms of various performance metrics, on nine open multi-label classification datasets.
When FALT is equipped with Mercer kernels, it is susceptible to \emph{curse of kernelization} \cite{WangCV12}, just like in the single-label case. Therefore, we plan to extend the kernelized FALT in the resource-limited constraint.

\section*{Acknowledgment} 
This project is supported by National Natural Science Foundation of China (Nos. 61906165) and Natural Science Foundation of the Jiangsu Higher Education Institutions of China (Nos. 19KJB520064).


\appendix
\section{Proof of Theorem 1}
\begin{proof}
According to Eq. (\ref{rule}), we have that 
\begin{align}
& ||\bm W_t - \bm U_*||_{F}^2 -  ||\bm W_{t+1} - \bm U_*||_{F}^2 \nonumber\\
=& \sum_{i=1}^{L+1} \big( ||\bm w_t^{(i)} - \bm u_*^{(i)}||^2  - ||\bm w_{t}^{(i)} -\eta \nabla_t^{(i)} - \bm u_*^{(i)}||^2 \big) \nonumber\\
=& -\eta^2 \sum_{i=1}^{L+1} ||\nabla_t^{(i)}||^2 + 2 \eta \sum_{i=1}^{L+1} (\nabla_t^{(i)})^\top (\bm w_t^{(i)} - \bm u_*^{(i)}) \nonumber\\
\geq &  -\eta^2 \sum_{i=1}^{L+1} ||\nabla_t^{(i)}||^2 + 2 \eta \bigl( f_t (\bm W_t) - f_t (\bm U_*) \bigr)
\label{mid}
\end{align}
where the last inequality is due to the convexity of $f_t (\bm W)$.
Next we start to bound $\sum_{i=1}^{L+1} ||\nabla_t^{(i)}||^2$.
Three cases are checked on:
\begin{enumerate}
\item 
When $|Y_t| \geq 1$ and $|\bar Y_t| \geq 1$, by plugging into the expression of $\nabla_t^{(i)}$, we can get
\begin{align*}
\sum_{i=1}^{L+1} ||\nabla_t^{(i)}||^2 &= 
\left[
\frac{a_t}{|Y_t|^2} \!+\! \frac{b_t}{|\bar Y_t|^2} \!+\! \bigl(\frac{a_t}{|Y_t|} \!-\! \frac{b_t}{|\bar Y_t|} \bigr)^2 
\right]  ||\bm x_t ||^2  \\
&\leq \left[\frac{a_t}{|Y_t|} + \frac{b_t}{|\bar Y_t|} + |\frac{a_t}{|Y_t|} - \frac{b_t}{|\bar Y_t|}|  \right]  ||\bm x_t ||^2  \\
& \leq 2  ||\bm x_t ||^2
\end{align*}

\item 
When $|Y_t| = 0$ and $|\bar Y_t| \geq 1$, we have  
\begin{align*}
\sum_{i=1}^{L+1} ||\nabla_t^{(i)}||^2 
&= \left[\frac{b_t}{|\bar Y_t|^2} + \bigl(\frac{b_t}{|\bar Y_t|} \bigr)^2 \right]  ||\bm x_t ||^2  \\
&\leq \left[ \frac{b_t}{|\bar Y_t|} + \frac{b_t}{|\bar Y_t|} \right]  ||\bm x_t ||^2
\leq 2 ||\bm x_t ||^2
\end{align*}

\item 
When $|\bar Y_t| = 0$ and $|Y_t| \geq 1$, we can similarly get the same conclusion.
\end{enumerate}

In summary, we get $\sum_{i=1}^{L+1} ||\nabla_t^{(i)}||^2 \leq 2 ||\bm x_t ||^2 \leq 2 R^2$.
Plugging the inequality back into (\ref{mid}) and re-arranging these terms, we get that
\begin{align*}
 f_t (\bm W_t) \!-\! f_t (\bm U_*)  \! \leq \!
\frac{1}{2 \eta} ( ||\bm W_t \!-\! \bm U_*||_{F}^2  \!-\!  ||\bm W_{t+1} \!-\! \bm U_*||_{F}^2 ) \!+\! \eta R^2. 
\end{align*}
Summing the above inequality over $t=1$ to $T$, we get that
\begin{align*}
&\sum_{t=1}^T f_t (\bm W_t) \!-\! f_t (\bm U_*)   \\
\leq \
& \frac{1}{2 \eta} \sum_{t=1}^T ( ||\bm W_t - \bm U_*||_{F}^2  -  ||\bm W_{t+1} - \bm U_*||_{F}^2 ) 
+  \eta R^2 T  \\
= \
&\frac{1}{2 \eta} ( ||\bm W_1 - \bm U_*||_{F}^2  -  ||\bm W_{T+1} - \bm U_*||_{F}^2) + \eta R^2 T \\
\leq \
& \frac{1}{2 \eta} ||\bm U_*||_{F}^2 + \eta R^2 T .
\end{align*}
\end{proof}

\section{Proof of Theorem 2}

\begin{proof}
The optimality of $\bm W_{t+1}$ for the problem at Step \ref{secondorder} in Algorithm \ref{alg3} implies that
for any $\bm U_* = [\bm u_*^{(1)}, \cdots,\bm u_*^{(L+1)} ] \in \R^{d \times (L+1)}$, we have
\begin{align*}
\sum_{i=1}^{L+1} \bigl( \frac{1}{\eta} \bm H_t^{(i)} (\bm w_{t+1}^{(i)} - \bm w_t^{(i)}) + \nabla_t^{(i)} \bigr)^\top 
(\bm u_*^{(i)} - \bm w_{t+1}^{(i)}) \geq 0.
\end{align*}

By application of the inequality, we can get
\begin{align*}
  & \sum_{i=1}^{L+1} (\nabla_t^{(i)})^\top (\bm w_t^{(i)} - \bm u_*^{(i)}) 	 \\
=& \sum_{i=1}^{L+1}  (\nabla_t^{(i)} + \frac{1}{\eta} \bm H_t^{(i)} (\bm w_{t+1}^{(i)} - \bm w_t^{(i)}) )^\top (\bm w_{t+1}^{(i)} -\bm u_*^{(i)} )\\
  &+ \frac{1}{\eta} \sum_{i=1}^{L+1}  (\bm H_t^{(i)} (\bm w_{t}^{(i)} - \bm w_{t+1}^{(i)}) )^\top (\bm w_{t+1}^{(i)} -\bm u_*^{(i)} ) 
    + \sum_{i=1}^{L+1}  (\nabla_t^{(i)})^\top (\bm w_t^{(i)} - \bm w_{t+1}^{(i)})          \\
\leq &  
\frac{1}{\eta} \sum_{i=1}^{L+1} (\bm H_t^{(i)} (\bm w_{t}^{(i)} - \bm w_{t+1}^{(i)}))^\top (\bm w_{t+1}^{(i)} -\bm u_*^{(i)}) 
 + \sum_{i=1}^{L+1} (\nabla_t^{(i)})^\top (\bm w_t^{(i)} - \bm w_{t+1}^{(i)})   \\                      
= & \frac{1}{2\eta}\! \sum_{i=1}^{L+1} \!\left\{\! \|\bm w_t^{(i)} \!-\! \bm u_*^{(i)} \|_{\bm H_t^{(i)}}^2 
     \!-\! \| \bm w_{t+1}^{(i)} \!-\! \bm u_*^{(i)}\|_{\bm H_t^{(i)}}^2 
     \!-\! \| \bm w_{t+1}^{(i)} \!-\! \bm w_t^{(i)}\|_{\bm H_t^{(i)}}^2 \!\right\}\! \\                                               
   &+ \sum_{i=1}^{L+1}  (\nabla_t^{(i)})^\top (\bm w_t^{(i)} - \bm w_{t+1}^{(i)}) \\
\leq &  \frac{1}{2\eta}\! \sum_{i=1}^{L+1} \!\left\{\! \|\bm w_t^{(i)} \!-\! \bm u_*^{(i)} \|_{\bm H_t^{(i)}}^2 
     \!-\! \| \bm w_{t+1}^{(i)} \!-\! \bm u_*^{(i)}\|_{\bm H_t^{(i)}}^2 \!\right\}\! 
     + \frac{\eta}{2} \sum_{i=1}^{L+1} \|\nabla_t^{(i)}\|_{(\bm H_t^{(i)})^{-1}}^2
\end{align*}
where the last inequality follows by application of Fenchel-Yong's inequality to the conjugate functions $\frac{1}{2}\|\cdot\|_{\bm H_t^{(i)}}^2$ and $\frac{1}{2}\|\cdot\|_{(\bm H_t^{(i)})^{-1}}^2$:
\begin{align*}   
     & (\nabla_t^{(i)})^\top (\bm w_t^{(i)} - \bm w_{t+1}^{(i)}) \\
\leq \ & \frac{1}{2} \|\sqrt{\eta}  \nabla_t^{(i)} \|_{(\bm H_t^{(i)})^{-1}}^2 
           +\frac{1}{2}\|\frac{1}{\sqrt{\eta}} (\bm w_t^{(i)} - \bm w_{t+1}^{(i)})\|_{\bm H_t^{(i)}}^2  \\
= \  & \frac{\eta}{2} \|\nabla_t^{(i)} \|_{(\bm H_t^{(i)})^{-1}}^2 
          +\frac{1}{2 \eta}\| \bm w_t^{(i)} - \bm w_{t+1}^{(i)}\|_{\bm H_t^{(i)}}^2.  
\end{align*}

According to the convexity of $f_t(\bm W)$, we can get
\begin{align*}
 f_t (\bm W_t) - f_t (\bm U_*) \leq \sum_{i=1}^{L+1} (\nabla_t^{(i)})^\top (\bm w_t^{(i)} - \bm u_*^{(i)}) .
\end{align*}
By combining this with the above equation, we can get 
\begin{align*}
 f_t (\bm W_t) \!-\! f_t (\bm U_*) \leq & 
 \frac{1}{2\eta}\! \sum_{i=1}^{L+1} \!\left\{\! \|\bm w_t^{(i)} \!-\! \bm u_*^{(i)} \|_{\bm H_t^{(i)}}^2 
  \!-\! \| \bm w_{t+1}^{(i)} \!-\! \bm u_*^{(i)}\|_{\bm H_t^{(i)}}^2 \!\right\}\! \\
 & + \frac{\eta}{2} \sum_{i=1}^{L+1} \|\nabla_t^{(i)}\|_{(\bm H_t^{(i)})^{-1}}^2 .
\end{align*}
Summing the inequality over $t=1, \cdots, T$, we get
\begin{align}
&\sum_{t=1}^T ( f_t (\bm W_t) \!-\! f_t (\bm U_*))  \nonumber\\
\leq & \frac{1}{2\eta} \sum_{t=1}^T \sum_{i=1}^{L+1} \!\left\{\! \|\bm w_t^{(i)} \!-\! \bm u_*^{(i)} \|_{\bm H_t^{(i)}}^2 
     \!-\! \| \bm w_{t+1}^{(i)} \!-\! \bm u_*^{(i)}\|_{\bm H_t^{(i)}}^2 \!\right\}\! \nonumber\\
   & + \frac{\eta}{2} \sum_{t=1}^T \sum_{i=1}^{L+1} \|\nabla_t^{(i)}\|_{(\bm H_t^{(i)})^{-1}}^2 .
\label{inter}  
\end{align}

We first bound the first term in the right-hand side of (\ref{inter}):
\begin{align*}
&\sum_{t=1}^T \left\{ \|\bm w_t^{(i)} - \bm u_*^{(i)} \|_{\bm H_t^{(i)}}^2 
	    - \| \bm w_{t+1}^{(i)} - \bm u_*^{(i)}\|_{\bm H_t^{(i)}}^2 \right\}  \\
= & \sum_{t=1}^T \left\{ \|\bm w_t^{(i)} - \bm u_*^{(i)} \|_{\diag (\bm s_t^{(i)})}^2 
	    - \| \bm w_{t+1}^{(i)} - \bm u_*^{(i)}\|_{\diag (\bm s_{t}^{(i)})}^2 \right\} \\
  & + \delta \sum_{t=1}^T \left\{ \|\bm w_t^{(i)} - \bm u_*^{(i)} \|_2^2 
	    - \| \bm w_{t+1}^{(i)} - \bm u_*^{(i)}\|_2^2 \right\}	    \\
= & \sum_{t=2}^T \bigl( \| \bm w_t^{(i)} - \bm u_*^{(i)} \|_{\diag(\bm s_t^{(i)})}^2  
    - \| \bm w_t^{(i)} - \bm u_*^{(i)} \|_{\diag(\bm s_{t-1}^{(i)})}^2  \bigr)   \\ 
  & + \| \bm w_1^{(i)} - \bm u_*^{(i)} \|_{\diag(\bm s_1^{(i)})}^2 - \| \bm w_{T+1}^{(i)} - \bm u_*^{(i)} \|_{\diag(\bm s_T^{(i)})}^2 \\
  & + \delta \bigl( \|\bm w_1^{(i)} - \bm u_*^{(i)} \|_2^2 - \| \bm w_{T+1}^{(i)} - \bm u_*^{(i)}\|_2^2 \bigr) \\
\leq &  \sum_{t=2}^T \| \bm w_t^{(i)} - \bm u_*^{(i)} \|_{\infty}^2 \Tr \bigl(\diag(\bm s_t^{(i)}) - \diag(\bm s_{t-1}^{(i)}) \bigr) \\
      & + \| \bm w_1^{(i)} - \bm u_*^{(i)} \|_{\infty}^2 \Tr \bigl(\diag(\bm s_1^{(i)}) \bigr) 
         + \delta \|\bm w_1^{(i)} - \bm u_*^{(i)} \|_2^2  \\
\leq & \max_{t \in [T]} \| \bm w_t^{(i)} - \bm u_*^{(i)}\|_{\infty}^2 \Tr \bigl(\diag(\bm s_T^{(i)}) \bigr) + \delta \|\bm u_*^{(i)} \|_2^2
\end{align*}

Next we start to bound the second term in the right-hand side of (\ref{inter}).
\begin{align*}
\sum_{t=1}^T  \|\nabla_t^{(i)}\|_{(\bm H_t^{(i)})^{-1}}^2 
& = \sum_{t=1}^T (\nabla_t^{(i)})^\top (\delta \bm{I} + \diag(\bm{s}_t^{(i)}))^{-1} \nabla_t^{(i)} \\
&\leq \sum_{t=1}^T (\nabla_t^{(i)})^\top (\diag(\bm{s}_t^{(i)}))^{-1} \nabla_t^{(i)} 
\leq_1  2 \sum_{j=1}^d \| \bm G_{1: T, j}^{(i)}\|_2  
\end{align*}	
where $\leq_1$ is owing to Lemma 4 in \cite{DuchiHS11}.

Plugging the above two equation into (\ref{inter}), we can get 
\begin{align*}
\sum_{t=1}^T ( f_t (\bm W_t) \!-\! f_t (\bm U_*) ) \leq 
& \frac{1}{2 \eta}\sum_{i=1}^{L+1} {\max_{t \in [T]} \|\bm w_t^{(i)} - \bm u_*^{(i)}\|_{\infty}^2} \Tr \bigl(\diag(\bm s_T^{(i)}) \bigr) \\
& + \frac{\delta}{2 \eta}\sum_{i=1}^{L+1}  \|\bm u_*^{(i)} \|_2^2 + \eta \sum_{i=1}^{L+1} \sum_{j=1}^d \| \bm G_{1: T, j}^{(i)}\|_2  \\
\leq & \frac{Q}{2 \eta}  \sum_{i=1}^{L+1} \Tr \bigl(\diag(\bm s_T^{(i)}) \bigr) + \eta \sum_{i=1}^{L+1} \sum_{j=1}^d \| \bm G_{1: T, j}^{(i)}\|_2 + \frac{\delta}{2 \eta} \|\bm U_*\|_F^2 .
\end{align*}
Using the fact that $\Tr \bigl(\diag(\bm s_T^{(i)}) \bigr)   = \sum_{j=1}^d \| \bm G_{1: T, j}^{(i)}\|_2$ concludes the proof.
\end{proof}

\bibliographystyle{elsarticle_num}
\bibliography{reference.bib}

\begin{thebibliography}{10}
\expandafter\ifx\csname url\endcsname\relax
  \def\url#1{\texttt{#1}}\fi
\expandafter\ifx\csname urlprefix\endcsname\relax\def\urlprefix{URL }\fi
\expandafter\ifx\csname href\endcsname\relax
  \def\href#1#2{#2} \def\path#1{#1}\fi

\bibitem{BoutellLSB04}
M.~R. Boutell, J.~Luo, X.~Shen, C.~M. Brown, Learning multi-label scene
  classification, Pattern Recognit. 37~(9) (2004) 1757--1771.

\bibitem{LiSL17}
Y.~Li, Y.~Song, J.~Luo, Improving pairwise ranking for multi-label image
  classification, in: {IEEE} Conference on Computer Vision and Pattern
  Recognition {(CVPR)}, 2017, pp. 1837--1845.

\bibitem{LiuC15}
S.~M. Liu, J.~Chen, A multi-label classification based approach for sentiment
  classification, Expert Syst. Appl. 42~(3) (2015) 1083--1093.

\bibitem{BurkhardtK18}
S.~Burkhardt, S.~Kramer, Online multi-label dependency topic models for text
  classification, Mach. Learn. 107~(5) (2018) 859--886.

\bibitem{ElisseeffW01}
A.~Elisseeff, J.~Weston, A kernel method for multi-labelled classification, in:
  Advances in Neural Information Processing Systems {(NIPS)}, {MIT} Press,
  2001, pp. 681--687.

\bibitem{Lin07}
R.-E. Fan, C.-J. Lin,
  \href{https://www.csie.ntu.edu.tw/~cjlin/papers/threshold.pdf}{A study on
  threshold selection for multi-label classification}, Tech. rep., National
  Taiwan University (2007).
\newline\urlprefix\url{https://www.csie.ntu.edu.tw/~cjlin/papers/threshold.pdf}

\bibitem{HariharanZVV10}
B.~Hariharan, L.~Zelnik{-}Manor, S.~V.~N. Vishwanathan, M.~Varma, Large scale
  max-margin multi-label classification with priors, in: Proceedings of the
  27th International Conference on Machine Learning {(ICML)}, Omnipress, 2010,
  pp. 423--430.

\bibitem{ZhangSK17}
L.~Zhang, S.~K. Shah, I.~A. Kakadiaris, Hierarchical multi-label classification
  using fully associative ensemble learning, Pattern Recognit. 70 (2017)
  89--103.

\bibitem{SiZKMDH17}
S.~Si, H.~Zhang, S.~S. Keerthi, D.~Mahajan, I.~S. Dhillon, C.~Hsieh, Gradient
  boosted decision trees for high dimensional sparse output, in: Proceedings of
  the 34th International Conference on Machine Learning {(ICML)}, Vol.~70,
  2017, pp. 3182--3190.

\bibitem{ZhuKZ18}
Y.~Zhu, J.~T. Kwok, Z.~Zhou, Multi-label learning with global and local label
  correlation, {IEEE} Trans. Knowl. Data Eng. 30~(6) (2018) 1081--1094.

\bibitem{ChenWWG19}
Z.~Chen, X.~Wei, P.~Wang, Y.~Guo, Multi-label image recognition with graph
  convolutional networks, in: {IEEE} Conference on Computer Vision and Pattern
  Recognition {(CVPR)}, 2019, pp. 5177--5186.

\bibitem{WangKWJ21}
R.~Wang, S.~Kwong, X.~Wang, Y.~Jia, Active \emph{k}-labelsets ensemble for
  multi-label classification, Pattern Recognit. 109 (2021) 107583.

\bibitem{AlotaibiF21}
R.~Alotaibi, P.~A. Flach, Multi-label thresholding for cost-sensitive
  classification, Neurocomputing 436 (2021) 232--247.

\bibitem{FurnkranzHMB08}
J.~F{\"{u}}rnkranz, E.~H{\"{u}}llermeier, E.~L. Menc{\'{\i}}a, K.~Brinker,
  Multilabel classification via calibrated label ranking, Mach. Learn. 73~(2)
  (2008) 133--153.

\bibitem{LargeronMG12}
C.~Largeron, C.~Moulin, M.~G{\'{e}}ry, Mcut: {A} thresholding strategy for
  multi-label classification, in: Proceedings of 11th International Symposium
  on Advances in Intelligent Data Analysis {(IDA)}, Vol. 7619 of Lecture Notes
  in Computer Science, Springer, 2012, pp. 172--183.

\bibitem{TangRN09}
L.~Tang, S.~Rajan, V.~K. Narayanan, Large scale multi-label classification via
  metalabeler, in: Proceedings of the 18th International Conference on World
  Wide Web, {(WWW)}, {ACM}, 2009, pp. 211--220.

\bibitem{VenkatesanEDPW17}
R.~Venkatesan, M.~J. Er, M.~Dave, M.~Pratama, S.~Wu, A novel online multi-label
  classifier for high-speed streaming data applications, Evol. Syst. 8~(4)
  (2017) 303--315.

\bibitem{DuV20}
J.~Du, C.~Vong, Robust online multilabel learning under dynamic changes in data
  distribution with labels, {IEEE} Trans. Cybern. 50~(1) (2020) 374--385.

\bibitem{ReadBHP12}
J.~Read, A.~Bifet, G.~Holmes, B.~Pfahringer, Scalable and efficient multi-label
  classification for evolving data streams, Mach. Learn. 88~(1-2) (2012)
  243--272.

\bibitem{NguyenNLNLS19}
T.~T. Nguyen, T.~T.~T. Nguyen, A.~V. Luong, Q.~V.~H. Nguyen, A.~W. Liew,
  B.~Stantic, Multi-label classification via label correlation and first order
  feature dependance in a data stream, Pattern Recognit. 90 (2019) 35--51.

\bibitem{NguyenDLLLM19}
T.~T. Nguyen, M.~T. Dang, A.~V. Luong, A.~W. Liew, T.~Liang, J.~McCall,
  Multi-label classification via incremental clustering on an evolving data
  stream, Pattern Recognit. 95 (2019) 96--113.

\bibitem{OsojnikPD17}
A.~Osojnik, P.~Panov, S.~Dzeroski, Multi-label classification via multi-target
  regression on data streams, Mach. Learn. 106~(6) (2017) 745--770.

\bibitem{OuyangG12}
H.~Ouyang, A.~G. Gray, Stochastic smoothing for nonsmooth minimizations:
  Accelerating {SGD} by exploiting structure, in: Proceedings of the 29th
  International Conference on Machine Learning {(ICML)}, 2012.

\bibitem{ParkC13}
S.~Park, S.~Choi, Online multi-label learning with accelerated nonsmooth
  stochastic gradient descent, in: {IEEE} International Conference on
  Acoustics, Speech and Signal Processing {(ICASSP)}, 2013, pp. 3322--3326.

\bibitem{GongYB20}
X.~Gong, D.~Yuan, W.~Bao, Online metric learning for multi-label
  classification, in: Proceedings of the 34th {AAAI} Conference on Artificial
  Intelligence {(AAAI)}, {AAAI} Press, 2020, pp. 4012--4019.

\bibitem{Shalev-Shwartz12}
S.~Shalev{-}Shwartz, Online learning and online convex optimization, Found.
  Trends Mach. Learn. 4~(2) (2012) 107--194.

\bibitem{Hoi2018}
S.~C.~H. Hoi, D.~Sahoo, J.~Lu, P.~Zhao, Online learning: {A} comprehensive
  survey, CoRR abs/1802.02871.

\bibitem{CrammerDKSS06}
K.~Crammer, O.~Dekel, J.~Keshet, S.~Shalev{-}Shwartz, Y.~Singer, Online
  passive-aggressive algorithms, J. Mach. Learn. Res. 7 (2006) 551--585.

\bibitem{Shalev-ShwartzSSC11}
S.~Shalev{-}Shwartz, Y.~Singer, N.~Srebro, A.~Cotter, Pegasos: primal estimated
  sub-gradient solver for {SVM}, Math. Program. 127~(1) (2011) 3--30.

\bibitem{CrammerDP12}
K.~Crammer, M.~Dredze, F.~Pereira, Confidence-weighted linear classification
  for text categorization, J. Mach. Learn. Res. 13 (2012) 1891--1926.

\bibitem{LuHWZL16}
J.~Lu, S.~C.~H. Hoi, J.~Wang, P.~Zhao, Z.~Liu, Large scale online kernel
  learning, J. Mach. Learn. Res. 17 (2016) 47:1--47:43.

\bibitem{DingMLCLNS18}
S.~Ding, B.~Mirza, Z.~Lin, J.~Cao, X.~Lai, T.~V. Nguyen, J.~Sepulveda, Kernel
  based online learning for imbalance multiclass classification, Neurocomputing
  277 (2018) 139--148.

\bibitem{ZhangZ07}
M.~Zhang, Z.~Zhou, {ML-KNN:} {A} lazy learning approach to multi-label
  learning, Pattern Recognit. 40~(7) (2007) 2038--2048.

\bibitem{ZhangZ14}
M.~Zhang, Z.~Zhou, A review on multi-label learning algorithms, {IEEE} Trans.
  Knowl. Data Eng. 26~(8) (2014) 1819--1837.

\bibitem{GibajaV15}
E.~Gibaja, S.~Ventura, A tutorial on multilabel learning, {ACM} Comput. Surv.
  47~(3) (2015) 52:1--52:38.

\bibitem{ZhangLLG18}
M.~Zhang, Y.~Li, X.~Liu, X.~Geng, Binary relevance for multi-label learning: an
  overview, Frontiers Comput. Sci. 12~(2) (2018) 191--202.

\bibitem{ReadPHF11}
J.~Read, B.~Pfahringer, G.~Holmes, E.~Frank, Classifier chains for multi-label
  classification, Mach. Learn. 85~(3) (2011) 333--359.

\bibitem{TeisseyreZS19}
P.~Teisseyre, D.~Zufferey, M.~Slomka, Cost-sensitive classifier chains:
  Selecting low-cost features in multi-label classification, Pattern Recognit.
  86 (2019) 290--319.

\bibitem{TsoumakasKV11}
G.~Tsoumakas, I.~Katakis, I.~P. Vlahavas, Random k-labelsets for multilabel
  classification, {IEEE} Trans. Knowl. Data Eng. 23~(7) (2011) 1079--1089.

\bibitem{PupoMV16}
O.~G.~R. Pupo, C.~Morell, S.~Ventura, Effective lazy learning algorithm based
  on a data gravitation model for multi-label learning, Inf. Sci. 340-341
  (2016) 159--174.

\bibitem{VensSSDB08}
C.~Vens, J.~Struyf, L.~Schietgat, S.~Dzeroski, H.~Blockeel, Decision trees for
  hierarchical multi-label classification, Mach. Learn. 73~(2) (2008) 185--214.

\bibitem{BiFFK20}
L.~Bi, D.~D. Feng, M.~J. Fulham, J.~Kim, Multi-label classification of
  multi-modality skin lesion via hyper-connected convolutional neural network,
  Pattern Recognit. 107 (2020) 107502.

\bibitem{Vapnik1998}
V.~Vapnik, Statistical learning theory, Wiley, 1998.

\bibitem{DuchiHS11}
J.~C. Duchi, E.~Hazan, Y.~Singer, Adaptive subgradient methods for online
  learning and stochastic optimization, J. Mach. Learn. Res. 12 (2011)
  2121--2159.

\bibitem{WangCV12}
Z.~Wang, K.~Crammer, S.~Vucetic, Breaking the curse of kernelization: budgeted
  stochastic gradient descent for large-scale {SVM} training, J. Mach. Learn.
  Res. 13 (2012) 3103--3131.

\end{thebibliography}

%
%
%
%

\end{document}